\documentclass[10pt,twoside,reqno]{article}
\usepackage{amsfonts}
\usepackage{graphicx}
\usepackage[dvips]{color}

\def\bkR{{\rm I\kern-.17em R}}
\def\bkC{{\rm ^{_|}\kern-.47em C}}

\begin{document}



\thispagestyle{empty}
\begin{center}
{\large{\textbf{Uncertainty Measures for Probabilistic Hesitant
Fuzzy Sets in Multiple Criteria Decision Making
}}}
\\
\vspace{0.5cm} {\bf B. Farhadinia}\footnote{Corresponding author.} \quad  {\bf U. Aickelin}$^\dagger$ \quad {\bf H.A. Khorshidi}$^\dagger$
 \vspace*{0cm}
\\
\small{Dept. Math., Quchan University of Technology, Iran.}\\
\small{\verb"bfarhadinia@qiet.ac.ir"}
\\
\small{$^\dagger$ Dept. Computing and Information Systems, University of Melbourne, Australia.}
\\
\small{\verb"uwe.aickelin@unimelb.edu.au"}
\\
\small{\verb"hadi.khorshidi@unimelb.edu.au"}
\end{center}

\newtheorem{Theorem}{\quad Theorem}[section]

\newtheorem{Proposition}{\quad Proposition}[section]

\newtheorem{Definition}[Theorem]{\quad Definition}

\newtheorem{Corollary}[Theorem]{\quad Corollary}

\newtheorem{Lemma}[Theorem]{\quad Lemma}

\newtheorem{Example}[Theorem]{\quad Example}

\newtheorem{Illustrative example}[Theorem]{\quad Illustrative example}

\newtheorem{Remark}[Theorem]{\quad Remark}

\newtheorem{Assumption}[Theorem]{\quad Assumption}

\newtheorem{Algorithm}{\quad Algorithm}[section]

\newtheorem{Counterexample}{\quad Counterexample}[section]

\newtheorem{Notice}{\quad Notice}[section]
\newtheorem{Note}{\quad Note}[section]
\newtheorem{Contraction}{\quad Contraction}[section]

\bigskip

\noindent {\bf Abstract.} 
This contribution reviews critically the existing entropy measures for probabilistic hesitant fuzzy sets (PHFSs), and demonstrates that these entropy measures fail to effectively distinguish a variety of different PHFSs in some cases. 
In the sequel, we develop a new axiomatic framework of entropy
measures for probabilistic hesitant fuzzy elements (PHFEs) by considering two facets of uncertainty associated with PHFEs which are known as fuzziness and non-specificity. 
Respect to each {kind} of uncertainty, 
{a number of formulae are derived  
to} permit flexible selection of PHFE entropy measures. {Moreover, based on the proposed PHFE entropy measures, we introduce some entropy-based distance measures which are used in the portion of
comparative analysis.}
Eventually, the proposed PHFE entropy measures {and PHFE entropy-based distance measures} are applied to 
decision making in the strategy initiatives where their reliability and effectiveness are verified.

\bigskip

\noindent{\bf{Keywords}}: Probabilistic hesitant fuzzy set, Entropy measure, Multiple criteria decision making.

%

%

\section{Introduction}

In real multiple criteria group decision making (MCGDM), we usually face to the situation in which the decision makers do not have the {same 
significance in the decision-making process.
For instance, we suppose that four decision makers are going to discuss the membership of an element to a set such that three decision makers assign $0.3$, and one decision maker would like to assign $0.7$. In such a situation, if we use the concept of hesitant fuzzy element (HFE) for describing the opinion of decision makers as $\{0.3,0.7\}$, then the preference of decision makers considered here as $0.7$ is lost. To overcome such a problem, we should consider the concept of probabilistic hesitant fuzzy set (PHFS) proposed firstly by  Zhu \cite{[30zhu]}.   
Indeed, the concept of PHFS is an extension of hesitant fuzzy set (HFS) \cite{[29tor]} by}
 taking the probability term into account for describing the preference information.
{In this setting, the above-mentioned opinion of decision makers is described by $\{0.3|{\frac{3}{4}},0.7|{\frac{1}{4}}\}$ which is nothing else than PHFS concept.}  
\textcolor{blue}{By the way, the study of decision making is an interdisciplinary enterprise involving political science, economics, and psychology, as well as philosophy and statistics. For instance,  
Yue et al. \cite{[31yue]} developed a number of probabilistic hesitant fuzzy aggregation operators to} tackle with the multiple criteria decision-making (MCDM) problems. 
\textcolor{blue}{Then, Zeng et al. \cite{[32zen]} unified the concepts of probability and ordered weighted averaging  to introduce the uncertain probabilistic ordered weighted averaging distance operator.} 
Furthermore, Ding et al. \cite{ding} introduced an interactive technique for those probabilistic hesitant fuzzy MCGDM problems encompassing incomplete weight information.
Zhang et al. \cite{[15anov]} proposed a basic way for ranking probabilistic hesitant fuzzy elements (PHFEs) on the basis of score and deviation degrees, and their comparison technique returns  the absolute priorities of PHFEs. In continue with that, Song et al. \cite{[anov]} overcame the deficiencies of Zhang et al.'s \cite{[15anov]} technique by developing a novel comparison
technique for PHFEs on the basis of possibility degree formula.

{It seems} that the PHFS theory is still at the initial
stage, and there exist some important aspects which have not been explored thoroughly such as PHFE entropy measure.
Generally, the concept of entropy \textcolor{blue}{\cite{fardet,farxu2,[Exm1]}} is used for measuring the uncertainty degree which is contained in a probability distribution.
So far, a lot of contributions have been made in work on entropy measures for HFSs, and many MCDM problems are available concerning on the hesitant fuzzy information. For instance, Farhadinia \textcolor{blue}{\cite{[5EntM],[38tfar]}} developed a number of distance-based
entropy measures for HFSs; and Wei et al. \cite{[21EntM]} proposed a variety of HFS entropy measures by combining the score
and deviation functions of HFSs for computing the
criteria weights in {an} MCDM problem. 
The above-mentioned entropy measures 
{do not consider the probabilistic view of HFSs.}
\\ 
{One of the rare study} dedicated to PHFS entropy measures is that provided by Su et al. \cite{su} in which 
the membership degree-based and the distance-based entropies for PHFEs have been developed. 
\textcolor{blue}{They proposed} a like-distance based entropy measure for PHFEs which is 
based on the expectation concept in probability theory. 
By the way, each of \textcolor{blue}{Su et al.'s} 
membership degree-based and distance-based entropies for PHFEs has {the shortcomings} which are 
{listed as the contributions of our study:}
\begin{description}
\item[\textcolor{blue}{(i)}] {In view of \textcolor{blue}{Su et al.'s 
membership} degree-based entropy measures (that is, the property $(EP4)$ in Definition 4 \textcolor{blue}{of} \cite{su}), we are not able to find out any information } about the situation in which the corresponding probabilities are different. In a more particular case, the first part of property $(EP4)$ is based on the equality relationship of the corresponding probabilities, and this condition is abandoned in the second part. In addition to this limitation, 
\textcolor{blue}{Su et al. 
admitted} that  the membership degree-based entropy measures are invalid whenever $h(x)|p(x)=\{0|0.5,1|0.5\}$. 
\item[\textcolor{blue}{(ii)}] \textcolor{blue}{Su et al. 
employed} the concept of like-distance measure in constructing the distance-based entropy measures for PHFEs. However, one of the {shortcomings} of PHFE like-distance measure is that its reduced form does not coincide with the traditional definition of
HFE distance measure.
\end{description}
These 
{shortcomings show the necessity of developing new PHFE entropy measures.} 
In this study, we propose new entropy measures based on 
\textcolor{blue}{the two facets of uncertainty which was first generalized by Pal et al. \cite{pal} for intuitionistic fuzzy sets.
These facets of uncertainty, one of
which is related to fuzziness while the other is related to non-specificity, are employed here to define  
the fuzziness and non-specificity aspects for PHFEs. However,}
{by taking the proposed PHFE entropy measures into account, we \textcolor{blue}{then} develop a number of PHFE entropy-based distance measures which help to increase the efficiency of comparative analysis.}
{\textcolor{blue}{In the sequel,} we compare the} proposed PHFE entropy measures with the existing ones {to
illustrate} the merit and  {use} of the
proposed ones in real-life  {decision-making} scenarios.

This manuscript is organized as follows: In Section 2, 
we discuss thoroughly on the fuzziness- and non-specificity-based entropy measures for PHFSs, 
{and furthermore, we introduce some entropy-based distance measures for PHFEs.}
Section 3 deals with
a comparative analysis of the proposed PHFE entropy measures together with investigating the entropy-based decision-making with PHFE information.
Section 4 concludes the contribution and gives a vision to the future work.



\section{{\textcolor{blue}{Preliminaries}}}


\textcolor{blue}{In this section, we give some preliminaries and basic definitions used in the subsequent discussions.}
\\
Following Torra \cite{[29tor]}'s  {introduction of initial}
definition of hesitant fuzzy set (HFS), Xia and Xu \cite{[6xia]}  {extended that definition with the following mathematical representation.}
\begin{Definition}\cite{[6xia]}\label{Definition:HFS}
Assume that $X$ is the reference set. Then, a hesitant fuzzy set (HFS) on $X$ is defined in terms of a function from $X$ to a subset of $[0, 1]$ which is characterized by 
\begin{eqnarray*}
A=\{\langle x,h_{A}(x) \rangle ~|~x \in X\},
\end{eqnarray*}
in which $h_{A}(x)$ denotes a set of some values in $[0,1]$, and is called a hesitant fuzzy element (HFE). 
\end{Definition}
In the case where we associate each element of HFE $h_{A}(x)$ with its probability occurrence, then we have:
\begin{Definition}\cite{[8xu]}\label{pHFS}
Assume that $X$ is the reference set. Then, a probabilistic hesitant fuzzy set (PHFS) on $X$ is defined in terms of a function from $X$ to a subset of $[0, 1]$ which is characterized by 
\begin{eqnarray*}
A=\{\langle x,h_{A}(x)|p_A(x) \rangle ~|~x \in X\},
\end{eqnarray*}
in which $h_{A}(x)$ and $p_{A}(x)$ denote the sets of some values in $[0,1]$, and $h_{A}(x)|p_A(x)$ is called a probabilistic hesitant fuzzy element (PHFE). Here,  
$h_{A}(x)$ and $p_{A}(x)$ indicate respectively the possible membership degrees of the element
$x\in X$ to the set $A$; and the probabilities associated with $h_{A}(x)$. Moreover, the probability elements of any PHFE $h_A(x)|p_A(x)=\{h^{\sigma(i)}_A(x)|p^{\sigma(i)}_A(x)\}_{i=1}^{l_{xA}}$ satisfies $\sum_{i=1}^{l_{xA}}p^{\sigma(i)}_A(x)=1$ and $0\leq p^{\sigma(i)}_A(x) \leq 1$ for any $i=1,2,...,l_{xA}$ {, and moreover, 
$h^{\sigma(i)}_A(x)$ (or $p^{\sigma(i)}_A(x)$) is referred to as the $i$-th largest value in $h_A(x)$ (or $p_A(x)$).}
\end{Definition}


\textcolor{blue}{
Here, we introduce {the notation of $\Pi$ which is inspired by the two notions of 
deviation (of diverging probabilities) and mean (of equal probabilities). Suppose that $h_A(x)|p_A(x)=\{h^{\sigma(i)}_A(x)|p^{\sigma(i)}_A(x)\}_{i=1}^{l_{xA}}$ denotes a PHFE. Then, we define}
\begin{eqnarray}\label{pi}
\Pi(p^{{\sigma(i)}}(x),p^{{\sigma(j)}}(x))=\left\{
  \begin{array}{ll}
   |p^{{\sigma(i)}}(x)-p^{{\sigma(j)}}(x)|, & \hbox{$p^{{\sigma(i)}}(x)\not=p^{{\sigma(j)}}(x)$,} \\
    \frac{p^{{\sigma(i)}}(x)+p^{{\sigma(j)}}(x)}{2}, & \hbox{$p^{{\sigma(i)}}(x)=p^{{\sigma(j)}}(x)$,} 
  \end{array}
\right.
\end{eqnarray}
where $\Pi\in (0,1]$  for any $p^{{\sigma(i)}}(x),p^{{\sigma(j)}}(x)\in (0,1]$.
}

\textcolor{blue}{In this contribution}, we are \textcolor{blue}{mainly} interested to establish a number of entropy measures for PHFSs by taking two kinds of uncertainty including fuzziness and non-specificity into account for coping with the challenges may be stored in the related application context.
\textcolor{blue}{Briefly stating,} the fuzziness concept of PHFEs comes from the departure of PHFE from its nearest crisp set, and the non-specificity concept of PHFEs describes the imprecise knowledge which is contained in those PHFEs.

\begin{Definition}\label{ent-zhao} Let $h_A(x)|p_A(x)=\{h^{\sigma(i)}_A(x)|p^{\sigma(i)}_A(x)\}_{i=1}^{l_{xA}}$ and $h_B(x)|p_B(x)=\{h^{\sigma(i)}_B(x)|p^{\sigma(i)}_B(x)\}_{i=1}^{l_{xB}}$ { {be}} two PHFEs on $X $. 
{{Then the pair $(Ep_F,Ep_{NS})$ is called a two-tuple entropy measure for PHFEs if it possesses the following properties:}}
\begin{description}
\item[($\textrm{\textbf{Ep}}_\textrm{\textbf{F}}$0)] $0\leq Ep_F(h_A(x)|p_A(x))\leq 1$;
  \item[($\textrm{\textbf{Ep}}_\textrm{\textbf{F}}$1)] $Ep_F(h_A(x)|p_A(x))=0$ if and only if $h_A(x)|p_A(x)=O^*|1$ or $h_A(x)|p_A(x)=I^*|1$;
  \item[($\textrm{\textbf{Ep}}_\textrm{\textbf{F}}$2)] $Ep_F(h_A(x)|p_A(x))=1$ if and only if $h_A(x)|p_A(x)=\{{\frac{1}{2}}|1\}$;
  \item[($\textrm{\textbf{Ep}}_\textrm{\textbf{F}}$3)] $Ep_F(h_A(x)|p_A(x))=Ep_F(h_{A^c}(x)|p_{A^c}(x))$ {where the notation $A^c$ indicates the complement 
  of $A$;}
  \item[($\textrm{\textbf{Ep}}_\textrm{\textbf{F}}$4)] If $h_A^{\sigma(j)}(x)\leq h_B^{\sigma(j)}(x)\leq \frac{1}{2}$ and 
  $\Pi(p_A^{{\sigma(i)}}(x),p_A^{{\sigma(j)}}(x))\leq \Pi(p_B^{{\sigma(i)}}(x),p_B^{{\sigma(j)}}(x))$ 
  or $h_A^{\sigma(j)}(x)\geq h_B^{\sigma(j)}(x)\geq \frac{1}{2}$ and 
  $\Pi(p_A^{{\sigma(i)}}(x),p_A^{{\sigma(j)}}(x))\geq \Pi(p_B^{{\sigma(i)}}(x),p_B^{{\sigma(j)}}(x))$, 
  then $Ep_F(h_A(x))\leq Ep_F(h_B(x))$;
\end{description}
and
\begin{description}
\item[($\textrm{\textbf{Ep}}_\textrm{\textbf{NS}}$0)] $0\leq Ep_{NS}(h_A(x)|p_A(x))\leq 1$;
  \item[($\textrm{\textbf{Ep}}_\textrm{\textbf{NS}}$1)] $Ep_{NS}(h_A(x)|p_A(x))=0$ if and only if $h_A(x)|p_A(x)$ is a singleton, i.e., $h_A(x)|p_A(x)=\{\gamma|1\}$;
  \item[($\textrm{\textbf{Ep}}_\textrm{\textbf{NS}}$2)] $Ep_{NS}(h_A(x)|p_A(x))=1$ if and only if $h_A(x)|p_A(x)=\{0|\frac{1}{2},1|\frac{1}{2}\}$;
  \item[($\textrm{\textbf{Ep}}_\textrm{\textbf{NS}}$3)] $Ep_{NS}(h_A(x)|p_A(x))=Ep_{NS}(h_{A^c}(x)|p_{A^c}(x))$;
  \item[($\textrm{\textbf{Ep}}_\textrm{\textbf{NS}}$4)] If $|h_A^{\sigma(i)}(x)-h_A^{\sigma(j)}(x)|< |h_B^{\sigma(i)}(x)-h_B^{\sigma(j)}(x)|$
  and 
  $\Pi(p_A^{{\sigma(i)}}(x),p_A^{{\sigma(j)}}(x))= \Pi(p_B^{{\sigma(i)}}(x),p_B^{{\sigma(j)}}(x))$; or 
   $|h_A^{\sigma(i)}(x)-h_A^{\sigma(j)}(x)|= |h_B^{\sigma(i)}(x)-h_B^{\sigma(j)}(x)|$
  and 
  $\Pi(p_A^{{\sigma(i)}}(x),p_A^{{\sigma(j)}}(x))> \Pi(p_B^{{\sigma(i)}}(x),p_B^{{\sigma(j)}}(x))$
   for any $i,j=1,2,...,l_x$, then $Ep_{NS}(h_A(x))\leq Ep_{NS}(h_B(x))$.
\end{description}
\end{Definition}
According to the above definition, one can find that the two-tuple entropy measure $(Ep_F,Ep_{NS})$ describes how far the PHFE is from its
closest crisp counterpart, and moreover, how non-specific is the information which is expressed by the PHFE.

\section{{\textcolor{blue}{Entropy measures for PHFSs}}}

\textcolor{blue}{As the main part of the contribution, this section is dedicated to introducing three types of entropy measures for PHFSs which are referred here to as  fuzziness-based entropy measures, non-specificity-based entropy measures, and comprehensive  entropy measures for PHFEs. Then, this section ends with a focus on the PHFE distance measures which are based on  entropy measures.}  

\subsection{{\textcolor{blue}{Fuzziness-based entropy measure for PHFEs}}}

In the following, we give a brief description of generating a class of \textcolor{blue}{entropy} measures that quantify the fuzziness of a PHFE: 
\begin{Theorem}
\label{uncert-T1} Let $\Pi$ be the operator defined by $(\ref{pi})$, and 
$R: [0, 1]^2\rightarrow [0, 1]$ be a mapping which satisfies the following properties:
\begin{description}
\item[(R1)] $R(x,y)=0$ if and only if $x=y=0$ or $x=y=1$;
  \item[(R2)] $R(x,y)=1$ if and only if $x=y=\frac{1}{2}$;
  \item[(R3)] $R(x,y)=R(1-y,1-x)$;
  \item[(R4)] If $0\leq x_1\leq x_2\leq \frac{1}{2}$, $0\leq y_1\leq y_2\leq \frac{1}{2}$, then $R(x_1,y_1)\leq R(x_2,y_2)$; and if
$\frac{1}{2}\leq x_1\leq x_2\leq 1$, $\frac{1}{2}\leq y_1\leq y_2\leq 1$, then $R(x_1,y_1)\geq R(x_2,y_2)$.
\end{description}
Then, the mapping $Ep_F$ defined by
\begin{eqnarray}\label{Ep-F}
Ep_F(h(x)|p(x))=\frac{2}{l_x(l_x+1)}\sum_{i=1}^{l_x}\sum_{j\geq i}R(h^{\sigma(i)}(x),h^{\sigma(j)}(x))\Pi(p^{{\sigma(i)}}(x),p^{{\sigma(j)}}(x))
\end{eqnarray}
fulfils the axioms ($\textrm{\textbf{Ep}}_\textrm{\textbf{F}}\textrm{0}$)-($\textrm{\textbf{Ep}}_\textrm{\textbf{F}}\textrm{4}$) in Definition \ref{ent-zhao}.
\end{Theorem}
Proof. The proof of  axiom ($\textrm{\textbf{Ep}}_\textrm{\textbf{F}}\textrm{0}$) is straightforward.
\\
Proof of axiom ($\textrm{\textbf{Ep}}_\textrm{\textbf{F}}\textrm{1}$): Taking $h_A(x)|p_A(x)=O^*|1$ or $h_A(x)|p_A(x)=I^*|1$ into account, we deduce that $R(0,0)=0$ and $\Pi(1,1)=1$ or 
$R(1,1)=0$ and $\Pi(1,1)=1$. Thus, 
\begin{eqnarray*}
Ep_F(O^*|1)=\frac{2}{1(1+1)}R(0,0)\Pi(1,1)=0,\\
Ep_F(I^*|1)=\frac{2}{1(1+1)}R(1,1)\Pi(1,1)=0.
\end{eqnarray*}
Conversely, if $Ep_F(h(x)|p(x))=\frac{2}{l_x(l_x+1)}\sum_{i=1}^{l_x}\sum_{j\geq i}R(h^{\sigma(i)}(x),h^{\sigma(j)}(x))\Pi(p^{{\sigma(i)}}(x),p^{{\sigma(j)}}(x))
=0$, then we conclude that it must be satisfied $R(h^{\sigma(i)}(x),h^{\sigma(j)}(x))=0$ or $\Pi(p^{{\sigma(i)}}(x),p^{{\sigma(j)}}(x))=0$ for any $i,j=1,2,...,l_x$. Since from definition of $\Pi$, we find that $0<\Pi(p^{{\sigma(i)}}(x),p^{{\sigma(j)}}(x))$ holds true for any $i,j=1,2,...,l_x$, therefore, 
we result that $R(h^{\sigma(i)}(x),h^{\sigma(j)}(x))=0$ for any $i,j=1,2,...,l_x$, that is, $h^{\sigma(i)}(x)=h^{\sigma(j)}(x)=0$ or $h^{\sigma(i)}(x)=h^{\sigma(j)}(x)=1$. On the other hand, since the repetition of $p^{{\sigma(i)}}(x)$ is not meaningful, that is, we cannot encounter $\{O^*|\frac{1}{2},O^*|\frac{1}{2}\} $ or $\{O^*|\frac{1}{3},O^*|\frac{1}{3},O^*|\frac{1}{3}\} $, hence, it must be consider only $O^*|1$ or also $I^*|1$. 
\\
Proof of axiom ($\textrm{\textbf{Ep}}_\textrm{\textbf{F}}\textrm{2}$): Consider the case where $h_A(x)|p_A(x)=\{{\frac{1}{2}}|1\}$, thus 
\begin{eqnarray*}
Ep_F(h(x)|p(x))=\frac{2}{1(1+1)}R(\frac{1}{2},\frac{1}{2})\Pi(1,1)=R(\frac{1}{2},\frac{1}{2})
\end{eqnarray*}
therefore, from the property (R2), it results that $Ep_F(h(x)|p(x))=1$.
\\
Conversely, if $Ep_F(h(x)|p(x))=\frac{2}{l_x(l_x+1)}\sum_{i=1}^{l_x}\sum_{j\geq i}R(h^{\sigma(i)}(x),h^{\sigma(j)}(x))\Pi(p^{{\sigma(i)}}(x),p^{{\sigma(j)}}(x))
=1$, then we conclude that it must be satisfied $R(h^{\sigma(i)}(x),h^{\sigma(j)}(x))\Pi(p^{{\sigma(i)}}(x),p^{{\sigma(j)}}(x))=1$ for any $i,j=1,2,...,l_x$, or equivalently, $R(h^{\sigma(i)}(x),h^{\sigma(j)}(x))=1$ and $\Pi(p^{{\sigma(i)}}(x),p^{{\sigma(j)}}(x))=1$. Thus, the first part of the latter axiom together with the property (R2) give $h^{\sigma(i)}(x)=h^{\sigma(j)}(x)=\frac{1}{2}$, and moreover, the second part of the axiom results that $p^{{\sigma(i)}}(x)=p^{{\sigma(j)}}(x)=1$. This clearly means that  $h_A(x)|p_A(x)=\{{\frac{1}{2}}|1\}$.
\\
Proof of axiom ($\textrm{\textbf{Ep}}_\textrm{\textbf{F}}\textrm{3}$): From definition of $h_{A^c}(x)|p_{A^c}(x)$, we conclude that
\begin{eqnarray*}
Ep_F(h_{A^c}(x)|p_{A^c}(x))&=&\frac{2}{l_x(l_x+1)}\sum_{i=1}^{l_x}\sum_{j\geq i}R(1-h^{\sigma(i)}(x),1-h^{\sigma(j)}(x))\Pi((p^{\sigma(i)}(x))^c,(p^{{\sigma(j)}}(x))^c)\\
&&=\frac{2}{l_x(l_x+1)}\sum_{i=1}^{l_x}\sum_{j\geq i}R(1-h^{\sigma(i)}(x),1-h^{\sigma(j)}(x))\Pi(p^{{\sigma(i)}}(x),p^{{\sigma(j)}}(x)),
\end{eqnarray*}
and from the property (R3), we deduce that
\begin{eqnarray*}
&&\frac{2}{l_x(l_x+1)}\sum_{i=1}^{l_x}\sum_{j\geq i}R(1-h^{\sigma(i)}(x),1-h^{\sigma(j)}(x))\Pi(p^{{\sigma(i)}}(x),p^{{\sigma(j)}}(x))\\
&&=\frac{2}{l_x(l_x+1)}\sum_{i=1}^{l_x}\sum_{j\geq i}R(h^{\sigma(i)}(x),h^{\sigma(j)}(x))\Pi(p^{{\sigma(i)}}(x),p^{{\sigma(j)}}(x))
\end{eqnarray*}
that is, $Ep_F(h_{A^c}(x)|p_{A^c}(x))=Ep_F(h_{A}(x)|p_{A}(x))$.
\\
Proof of axiom ($\textrm{\textbf{Ep}}_\textrm{\textbf{F}}\textrm{4}$): 
Suppose that $h_A^{\sigma(j)}(x)\leq h_B^{\sigma(j)}(x)\leq \frac{1}{2}$ and 
  $\Pi(p_A^{{\sigma(i)}}(x),p_A^{{\sigma(j)}}(x))\leq \Pi(p_B^{{\sigma(i)}}(x),p_B^{{\sigma(j)}}(x))$ are to be held for any $i,j=1,2,...,l_x$. Then, from the property (R4) and the first aim, we get  $R(h_A^{\sigma(i)}(x),h_A^{\sigma(j)}(x))\leq R(h_B^{\sigma(i)}(x),h_B^{\sigma(j)}(x))$.
  Therefore, 
\begin{eqnarray*}
Ep_F(h_{A}(x)|p_{A}(x))&=&\frac{2}{l_x(l_x+1)}\sum_{i=1}^{l_x}\sum_{j\geq i}R(h_A^{\sigma(i)}(x),h_A^{\sigma(j)}(x))\Pi(p_A^{{\sigma(i)}}(x),p_A^{{\sigma(j)}}(x))\\
&\leq &\frac{2}{l_x(l_x+1)}\sum_{i=1}^{l_x}\sum_{j\geq i}R(h_B^{\sigma(i)}(x),h_B^{\sigma(j)}(x))\Pi(p_B^{{\sigma(i)}}(x),p_B^{{\sigma(j)}}(x))=
Ep_F(h_{B}(x)|p_{B}(x)).
\end{eqnarray*}  
The other case is proven similarly. $\Box$

By a simple investigation, we observe from Theorem \ref{uncert-T1} that
\begin{eqnarray*}
Ep_F(\{0|\frac{1}{2},1|\frac{1}{2}\})&=&\frac{1}{3}[R(0,0)\Pi(\frac{1}{2},\frac{1}{2})+R(0,1)\Pi(\frac{1}{2},\frac{1}{2})+R(1,1)\Pi(\frac{1}{2},\frac{1}{2})]\\
&=&
\frac{1}{3}[0+R(0,1)\frac{1}{2}+0]
=\frac{1}{6}R(0,1)\not=0(=R(0,0)=R(1,1)).
\end{eqnarray*}
This implies that the fuzzy entropy of the 
PHFE $h_A(x)|p_A(x)=\{0|\frac{1}{2},1|\frac{1}{2}\}$
 differs from those of the 
PHFE $h_A(x)|p_A(x)=O^*|1$ and the PHFE $h_A(x)|p_A(x)=I^*|1$,
which seems quite reasonable.

It is interesting to note that we are now able to develop the calculation of entropy $Ep_F$ by the help of following 
different formulas:
\begin{eqnarray*}
&&R_1(x,y)=[1-(\frac{1}{3}|1-4xy|)^r][1-(\frac{1}{3}|4(x+y-xy)-3|)^r],\quad r\geq 1;\\
&&R_2(x,y)=[\frac{2}{3}(\min\{1-2xy,xy\}+1)][\frac{2}{3}(\min\{2(x+y-xy)-1,2-2(x+y-xy)\}+1)];
\end{eqnarray*}
which 
generate respectively the fuzzy entropy formulas as:
\begin{eqnarray}\label{E-F2fi11}
&&Ep_{1F}(h(x))=\nonumber \\&&\frac{2}{l_x(l_x+1)}\sum_{i=1}^{l_x}\sum_{j\geq i}
[1-(\frac{1}{3}|1-4h^{\sigma(i)}(x)h^{\sigma(j)}(x)|)^r]\nonumber \\&&
\times[1-(\frac{1}{3}|4(h^{\sigma(i)}(x)+h^{\sigma(j)}(x)-h^{\sigma(i)}(x)h^{\sigma(j)}(x))-3|)^r]
\times \Pi(p_A^{{\sigma(i)}}(x),p_A^{{\sigma(j)}}(x)),\quad r\geq 1;\\
&&Ep_{2F}(h(x))=\nonumber \\&&\frac{2}{l_x(l_x+1)}\sum_{i=1}^{l_x}\sum_{j\geq i}
\frac{2}{3}(\min\{1-2h^{\sigma(i)}(x)h^{\sigma(j)}(x),h^{\sigma(i)}(x)h^{\sigma(j)}(x)\}+1)
\nonumber\\&&
\times
\frac{2}{3}(\min\{2(h^{\sigma(i)}(x)+h^{\sigma(j)}(x)-h^{\sigma(i)}(x)h^{\sigma(j)}(x))-1,\nonumber\\&&
\hspace{4cm} 2-2(h^{\sigma(i)}(x)+h^{\sigma(j)}(x)-h^{\sigma(i)}(x)h^{\sigma(j)}(x))\}+1)
\times \Pi(p_A^{{\sigma(i)}}(x),p_A^{{\sigma(j)}}(x)).\label{E-F2fi12}
\end{eqnarray}

\subsection{{\textcolor{blue}{Non-specificity-based entropy measure for PHFEs}}}

Here, let us discuss  the other aspect of uncertainty associated with PHFEs being referred to as non-specificity.\\
\begin{Theorem}\label{uncert-T1NS} Let $\Pi$ be the operator defined by $(\ref{pi})$, and $F: [0, 1]^2\rightarrow [0, 1]$ be a mapping which satisfies the following properties:
\begin{description}
\item[(F1)] $F(x,y)=0$ if and only if $x=y$;
  \item[(F2)] $F(x,y)=1$ if and only if $\{x,y\}\cap\{0,1\}\not =\emptyset$;
  \item[(F3)] $F(x,y)=F(1-y,1-x)$;
  \item[(F4)] $F(x_1,x_2)\geq F(x_3,x_4)$ holds true if $|x_1-x_2|\geq |x_3-x_4|$ where $x_i\in [0,1]$ for $i=1,2,3,4$.
\end{description}
Then, the mapping $Ep_{NS}$ defined by
\begin{eqnarray}\label{Ep-NS}
Ep_{NS}(h_A(x)|p_A(x))=\frac{2}{\max\{2,l_x(l_x-1)\}}\sum_{i=1}^{l_x}\sum_{j\geq i}F(h^{\sigma(i)}(x),h^{\sigma(j)}(x))
^{\Pi(p^{{\sigma(i)}}(x),p^{{\sigma(j)}}(x))}
\end{eqnarray}
fulfils the axioms ($\textrm{\textbf{Ep}}_\textrm{\textbf{NS}}\textrm{0}$)-($\textrm{\textbf{Ep}}_\textrm{\textbf{NS}}\textrm{4}$) in Definition \ref{ent-zhao}.
\end{Theorem}
Proof. The proof of  axiom ($\textrm{\textbf{Ep}}_\textrm{\textbf{NS}}\textrm{0}$) is straightforward.
\\
Proof of axiom ($\textrm{\textbf{Ep}}_\textrm{\textbf{NS}}\textrm{1}$): If $h_A(x)|p_A(x)$ is a singleton, i.e., $h_A(x)|p_A(x)=\{\gamma|1\}$, then by considering the definition of $Ep_{NS}$ given by (\ref{Ep-NS}), we find that
$Ep_{NS}(h_A(x)|p_A(x))=F(\gamma,\gamma)^{\Pi(1,1)}$. From the property (F1) and the fact that $\Pi(1,1)=1$, it results that $Ep_{NS}(h_A(x)|p_A(x))=0$.
\\
Conversely, if 
\begin{eqnarray*}
Ep_{NS}(h_A(x)|p_A(x))=\frac{2}{\max\{2,l_x(l_x-1)\}}\sum_{i=1}^{l_x}\sum_{j\geq i}F(h^{\sigma(i)}(x),h^{\sigma(j)}(x))^{\Pi(p^{{\sigma(i)}}(x),p^{{\sigma(j)}}(x))}=0,
\end{eqnarray*}
then we conclude that it should be satisfied $F(h^{\sigma(i)}(x),h^{\sigma(j)}(x))=0$ for any $i,j=1,2,...,l_x$. Hence, keeping the property (F2) in the mind, we get $h^{\sigma(i)}(x)=h^{\sigma(j)}(x)=\gamma$ for any $i,j=1,2,...,l_x$.
On the other hand, since the repetition of $p^{{\sigma(i)}}(x)$ is not meaningful, that is, we cannot encounter $\{\gamma|\frac{1}{2},\gamma|\frac{1}{2}\} $ or $\{\gamma|\frac{1}{3},\gamma|\frac{1}{3},\gamma|\frac{1}{3}\} $, hence, it must be consider only $\{\gamma|1\}$. 
\\
Proof of axiom ($\textrm{\textbf{Ep}}_\textrm{\textbf{NS}}\textrm{2}$):
Consider the case where $h_A(x)|p_A(x)=\{0|\frac{1}{2},1|\frac{1}{2}\}$, thus 
\begin{eqnarray*}
Ep_{NS}(h_A(x)|p_A(x))&=&\frac{2}{\max\{2,2(2-1)\}}[F(0,0)^{\Pi(\frac{1}{2},\frac{1}{2})}+F(0,1)^{\Pi(\frac{1}{2},\frac{1}{2})}+F(1,1)^{\Pi(\frac{1}{2},\frac{1}{2})}]
\end{eqnarray*}
therefore, from the properties (F2) and (F3), it results that $Ep_{NS}(h_A(x)|p_A(x))=F(0,1)^{\Pi(\frac{1}{2},\frac{1}{2})}=1$.
\\
Conversely, if $Ep_{NS}(h_A(x)|p_A(x))=\frac{2}{\max\{2,l_x(l_x-1)\}}\sum_{i=1}^{l_x}\sum_{j\geq i}F(h^{\sigma(i)}(x),h^{\sigma(j)}(x))^{\Pi(p^{{\sigma(i)}}(x),p^{{\sigma(j)}}(x))}=1$ and assume that $l_x=2$. Then, we get 
\begin{eqnarray*}
1&=&Ep_{NS}(h_A(x)|p_A(x))\\&=&\frac{2}{\max\{2,2(2-1)\}}
[F(h^{\sigma(1)}(x),h^{\sigma(1)}(x))^{\Pi(p^{{\sigma(1)}}(x),p^{{\sigma(1)}}(x))}+F(h^{\sigma(1)}(x),h^{\sigma(2)}(x))^{\Pi(p^{{\sigma(1)}}(x),p^{{\sigma(2)}}(x))}\\&&+F(h^{\sigma(2)}(x),h^{\sigma(2)}(x))^{\Pi(p^{{\sigma(2)}}(x),p^{{\sigma(2)}}(x))}]
\\
&=&
[0+F(h^{\sigma(1)}(x),h^{\sigma(2)}(x))^{\Pi(p^{{\sigma(1)}}(x),p^{{\sigma(2)}}(x))}+0]= F(h^{\sigma(1)}(x),h^{\sigma(2)}(x))^{\Pi(p^{{\sigma(1)}}(x),p^{{\sigma(2)}}(x))}.
\end{eqnarray*}
Since $0<\Pi(p^{{\sigma(1)}}(x),p^{{\sigma(2)}}(x))$, thus the above relation does not satisfy unless 
\begin{eqnarray*}
F(h^{\sigma(1)}(x),h^{\sigma(2)}(x))=1, 
\end{eqnarray*}
By taking the property (R2) into account, this means that $\{h^{\sigma(1)}(x),h^{\sigma(2)}(x))\}\cap\{0,1\}\not =\emptyset$.
On the other hand, since the repetition of $p^{{\sigma(i)}}(x)$ is not meaningful, that is, we cannot encounter $\{0|\frac{1}{3},0|\frac{1}{3},1|\frac{1}{3}\}$ or $\{0|\frac{1}{3},1|\frac{1}{3},1|\frac{1}{3}\}$, hence, it must be considered only $\{0|\frac{1}{2},1|\frac{1}{2}\}$, that is, 
\begin{eqnarray*}
 \Pi(p^{{\sigma(1)}}(x),p^{{\sigma(2)}}(x))=\frac{1}{2}.
\end{eqnarray*}
This means that it should be defined $h_A(x)|p_A(x)=\{0|\frac{1}{2},1|\frac{1}{2}\}$.
\\
Now we assume that $l_x=3$. Then, one gets 
\begin{eqnarray*}
1&=&Ep_{NS}(h_A(x)|p_A(x))=\frac{2}{\max\{2,3(3-1)\}}\\&&
[F(h^{\sigma(1)}(x),h^{\sigma(1)}(x))^{\Pi(p^{{\sigma(1)}}(x),p^{{\sigma(1)}}(x))}+F(h^{\sigma(1)}(x),h^{\sigma(2)}(x))^{\Pi(p^{{\sigma(1)}}(x),p^{{\sigma(2)}}(x))}\\&&\hspace{5 cm}+F(h^{\sigma(1)}(x),h^{\sigma(3)}(x))^{\Pi(p^{{\sigma(1)}}(x),p^{{\sigma(3)}}(x))}
\\&&+F(h^{\sigma(2)}(x),h^{\sigma(2)}(x))^{\Pi(p^{{\sigma(2)}}(x),p^{{\sigma(2)}}(x))}+F(h^{\sigma(2)}(x),h^{\sigma(3)}(x))^{\Pi(p^{{\sigma(2)}}(x),p^{{\sigma(3)}}(x))}
\\&&+F(h^{\sigma(3)}(x),h^{\sigma(3)}(x))^{\Pi(p^{{\sigma(3)}}(x),p^{{\sigma(3)}}(x))}
\\
&=&\frac{1}{3}\times
[0+F(h^{\sigma(1)}(x),h^{\sigma(2)}(x))^{\Pi(p^{{\sigma(1)}}(x),p^{{\sigma(2)}}(x))}+F(h^{\sigma(1)}(x),h^{\sigma(3)}(x))^{\Pi(p^{{\sigma(1)}}(x),p^{{\sigma(3)}}(x))}+0\\&&+F(h^{\sigma(2)}(x),h^{\sigma(3)}(x))^{\Pi(p^{{\sigma(2)}}(x),p^{{\sigma(3)}}(x))}
+0]=\\&&\frac{1}{3}\times [F(h^{\sigma(1)}(x),h^{\sigma(2)}(x))^{\Pi(p^{{\sigma(1)}}(x),p^{{\sigma(2)}}(x))}+
F(h^{\sigma(1)}(x),h^{\sigma(3)}(x))^{\Pi(p^{{\sigma(1)}}(x),p^{{\sigma(3)}}(x))}\\&&+F(h^{\sigma(2)}(x),h^{\sigma(3)}(x))^{\Pi(p^{{\sigma(2)}}(x),p^{{\sigma(3)}}(x))}].
\end{eqnarray*}
The above relation does not satisfy unless 
\begin{eqnarray*}
&&F(h^{\sigma(1)}(x),h^{\sigma(2)}(x))=1, \\
&&F(h^{\sigma(1)}(x),h^{\sigma(3)}(x))=1,\\
&&F(h^{\sigma(2)}(x),h^{\sigma(3)}(x))=1.
\end{eqnarray*}
Keeping the first two relations in the mind, we find that  
$h^{\sigma(1)}(x)=0,h^{\sigma(2)}(x))=1,h^{\sigma(3)}(x)=1$, and the third relation results itself in $h^{\sigma(2)}(x)=0,h^{\sigma(3)}(x))=1$ which is indeed contradictory, 
and therefore we need not to investigate any relation between 
$\Pi(p^{{\sigma(1)}}(x),p^{{\sigma(2)}}(x))$, $\Pi(p^{{\sigma(1)}}(x),p^{{\sigma(3)}}(x))$ and $\Pi(p^{{\sigma(2)}}(x),p^{{\sigma(3)}}(x))$.\\
Similar result is obtainable when we consider $l_x\geq 3$. 
\\
Proof of axiom ($\textrm{\textbf{Ep}}_\textrm{\textbf{NS}}\textrm{3}$): From definition of $h_{A^c}(x)|p_{A^c}(x)$, we conclude that
\begin{eqnarray*}
Ep_{NS}(h_{A^c}(x)|p_{A^c}(x))&=&\frac{2}{\max\{2,l_x(l_x-1)\}}\sum_{i=1}^{l_x}\sum_{j\geq i}F((h^{\sigma(i)}(x))^c,(h^{\sigma(j)}(x))^c)^{\Pi((p^{{\sigma(i)}}(x))^c,(p^{{\sigma(j)}}(x))^c)}\\
&=&\frac{2}{\max\{2,l_x(l_x-1)\}}\sum_{i=1}^{l_x}\sum_{j\geq i}F(1-h^{\sigma(i)}(x),1-h^{\sigma(j)}(x))^{\Pi(p^{{\sigma(i)}}(x),p^{{\sigma(j)}}(x))}
\end{eqnarray*}
and from the property (F3), we deduce that
\begin{eqnarray*}
Ep_{NS}(h_{A^c}(x)|p_{A^c}(x))=\frac{2}{\max\{2,l_x(l_x-1)\}}\sum_{i=1}^{l_x}\sum_{j\geq i}F(h^{\sigma(i)}(x),h^{\sigma(j)}(x))^{\Pi(p^{{\sigma(i)}}(x),p^{{\sigma(j)}}(x))}=
Ep_{NS}(h_A(x)|p_A(x)).
\end{eqnarray*}
\\
Proof of axiom ($\textrm{\textbf{Ep}}_\textrm{\textbf{NS}}\textrm{4}$): 
 If $|h_A^{\sigma(i)}(x)-h_A^{\sigma(j)}(x)|< |h_B^{\sigma(i)}(x)-h_B^{\sigma(j)}(x)|$
  and 
  $\Pi(p_A^{{\sigma(i)}}(x),p_A^{{\sigma(j)}}(x))= \Pi(p_B^{{\sigma(i)}}(x),p_B^{{\sigma(j)}}(x))$; or 
   $|h_A^{\sigma(i)}(x)-h_A^{\sigma(j)}(x)|= |h_B^{\sigma(i)}(x)-h_B^{\sigma(j)}(x)|$
  and 
  $\Pi(p_A^{{\sigma(i)}}(x),p_A^{{\sigma(j)}}(x))> \Pi(p_B^{{\sigma(i)}}(x),p_B^{{\sigma(j)}}(x))$
   for any $i,j=1,2,...,l_x$, then from the property (F4), we obviously conclude that $Ep_{NS}(h_A(x))\leq Ep_{NS}(h_B(x))$. $\Box$

Taking the above theorem into account, we are able to develop the calculation of entropy $Ep_{NS}$ by the use of following 
different formulas: 
\begin{eqnarray*}
&&F_1(x,y)=\frac{2|x-y|}{1+|x-y|};\\
&&F_2(x,y)=\frac{log(1+|x-y|)}{log(2)};\\
&&F_3(x,y)=|x-y|e^{|x-y|-1},
\end{eqnarray*}
which 
generate respectively  the entropy formulas  as:
\begin{eqnarray}\label{E-F2fi1NS}
&&Ep_{1NS}(h(x))=\frac{2}{\max\{2,l_x(l_x-1)\}}\sum_{i=1}^{l_x}\sum_{j\geq i}
[\frac{2|h^{\sigma(i)}(x)-h^{\sigma(j)}(x)|}{1+|h^{\sigma(i)}(x)-h^{\sigma(j)}(x)|}]^{\Pi(p_A^{{\sigma(i)}}(x),p_A^{{\sigma(j)}}(x))}; \\
&&Ep_{2NS}(h(x))=\frac{2}{\max\{2,l_x(l_x-1)\}}\sum_{i=1}^{l_x}\sum_{j\geq i}
[\frac{log(1+|h^{\sigma(i)}(x)-h^{\sigma(j)}(x)|)}{log(2)}]^{\Pi(p_A^{{\sigma(i)}}(x),p_A^{{\sigma(j)}}(x))};
\\
&&Ep_{3NS}(h(x))=\frac{2}{\max\{2,l_x(l_x-1)\}}\sum_{i=1}^{l_x}\sum_{j\geq i}
[|h^{\sigma(i)}(x)-h^{\sigma(j)}(x)|e^{|h^{\sigma(i)}(x)-h^{\sigma(j)}(x)|-1}]^{\Pi(p_A^{{\sigma(i)}}(x),p_A^{{\sigma(j)}}(x))} .\nonumber \\&&\label{E-F2fi2NS}
\end{eqnarray}

\subsection{{\textcolor{blue}{Comprehensive entropy measure for PHFEs}}}

In what follows, we are interested in presenting a unification framework of the two above-mentioned entropy measures $Ep_{NS}$ and $Ep_{F}$ for achieving a comprehensive entropy. 

\begin{Definition}\label{comp-ent} Let $h_A(x)|p_A(x)=\{h^{\sigma(i)}_A(x)|p^{\sigma(i)}_A(x)\}_{i=1}^{l_{xA}}$ and $h_B(x)|p_B(x)=\{h^{\sigma(i)}_B(x)|p^{\sigma(i)}_B(x)\}_{i=1}^{l_{xB}}$ { {be}} two PHFEs on $X$.
Then,  the mapping $Ep_{C}$ is called a comprehensive entropy measure for PHFEs  
if it possesses the following properties:
\begin{description}
\item[($\textrm{\textbf{Ep}}_\textrm{\textbf{C}}$0)] $0\leq Ep_C(h_A(x)|p_A(x))\leq 1$;
  \item[($\textrm{\textbf{Ep}}_\textrm{\textbf{C}}$1)] $Ep_C(h_A(x)|p_A(x))=0$ if and only if $h_A(x)|p_A(x)=O^*|1$ or $h_A(x)|p_A(x)=I^*|1$;
  \item[($\textrm{\textbf{Ep}}_\textrm{\textbf{C}}$2)] $Ep_C(h_A(x)|p_A(x))=1$ if and only if $h_A(x)|p_A(x)=\{{\frac{1}{2}}|1\}$;
  \item[($\textrm{\textbf{Ep}}_\textrm{\textbf{C}}$3)] $Ep_C(h_A(x)|p_A(x))=E_C(h_{A^c}(x)|p_{A^c}(x))$;
  \item[($\textrm{\textbf{Ep}}_\textrm{\textbf{C}}$4)] If $h_A^{\sigma(j)}(x)\leq h_B^{\sigma(j)}(x)\leq \frac{1}{2}$ and 
  $\Pi(p_A^{{\sigma(i)}}(x),p_A^{{\sigma(j)}}(x))\leq \Pi(p_B^{{\sigma(i)}}(x),p_B^{{\sigma(j)}}(x))$ 
  or $h_A^{\sigma(j)}(x)\geq h_B^{\sigma(j)}(x)\geq \frac{1}{2}$ and 
  $\Pi(p_A^{{\sigma(i)}}(x),p_A^{{\sigma(j)}}(x))\geq \Pi(p_B^{{\sigma(i)}}(x),p_B^{{\sigma(j)}}(x))$,
   together with \\
   if $|h_A^{\sigma(i)}(x)-h_A^{\sigma(j)}(x)|< |h_B^{\sigma(i)}(x)-h_B^{\sigma(j)}(x)|$
  and 
  $\Pi(p_A^{{\sigma(i)}}(x),p_A^{{\sigma(j)}}(x))= \Pi(p_B^{{\sigma(i)}}(x),p_B^{{\sigma(j)}}(x))$; or 
   $|h_A^{\sigma(i)}(x)-h_A^{\sigma(j)}(x)|= |h_B^{\sigma(i)}(x)-h_B^{\sigma(j)}(x)|$
  and 
  $\Pi(p_A^{{\sigma(i)}}(x),p_A^{{\sigma(j)}}(x))> \Pi(p_B^{{\sigma(i)}}(x),p_B^{{\sigma(j)}}(x))$
   for any $i,j=1,2,...,l_x$,\\ then  $Ep_C(h_A(x))\leq Ep_C(h_B(x))$.
\end{description}
\end{Definition}

Now, we assume that $\Theta:[0,1]^2\rightarrow [0,1]$ is a mapping which satisfies 
\begin{description}
\item [($\Theta0$)] $\Theta(x,0)=x$ for $x\in\{0,1\}$ (boundary condition);
\item [($\Theta1$)] $\Theta(x,y)=\Theta(y,x)$  (commutativity condition);
\item [($\Theta2$)] $\Theta(x,y)\leq \Theta(x,z)$ for any $y\leq z$ (monotonicity condition);
\end{description}
then, we conclude that:
\begin{Theorem}\label{comp-ent2} Let $Ep_{F}$ and $Ep_{NS}$ be those entropy measures given respectively by (\ref{Ep-F}) and (\ref{Ep-NS}). Then,  the mapping 
\begin{eqnarray}\label{Ep-C1}
Ep_C(h_A(x)|p_A(x))=\Theta(Ep_F(h_A(x)|p_A(x)),Ep_{NS}(h_A(x)|p_A(x)))
\end{eqnarray}
defines a comprehensive entropy measure for PHFEs.  
\end{Theorem}
Proof. It is sufficient to prove that the properties ($\textrm{\textbf{Ep}}_\textrm{\textbf{C}}$0)-($\textrm{\textbf{Ep}}_\textrm{\textbf{C}}$4) are to be held in Definition \ref{comp-ent}.
\\
From definition of $\Theta$, the proof of $(\textrm{\textbf{Ep}}_\textrm{\textbf{C}}0)$ is straightforward.
\\
Proof of axiom ($\textrm{\textbf{Ep}}_\textrm{\textbf{C}}\textrm{1}$): $Ep_C(h_A(x)|p_A(x))=0$ if and only if $\Theta(Ep_F(h_A(x)|p_A(x)),Ep_{NS}(h_A(x)|p_A(x)))=0$, that is, from the property ($\Theta0$), we conclude that it should be $Ep_F(h_A(x)|p_A(x))=0$ and $Ep_{NS}(h_A(x)|p_A(x))=0$. From the properties $(\textrm{\textbf{Ep}}_\textrm{\textbf{F}}1)$  and $(\textrm{\textbf{Ep}}_\textrm{\textbf{NS}}1)$, we result that it should be $h_A(x)|p_A(x)=O^*|1$ or $h_A(x)|p_A(x)=I^*|1$.
\\
Proof of axiom ($\textrm{\textbf{Ep}}_\textrm{\textbf{C}}\textrm{2}$):  $Ep_C(h_A(x)|p_A(x))=1$ if and only if $\Theta(Ep_F(h_A(x)|p_A(x)),Ep_{NS}(h_A(x)|p_A(x)))=1$, that is, from the property ($\Theta0$), we conclude that it should be $Ep_F(h_A(x)|p_A(x))=1$ and $Ep_{NS}(h_A(x)|p_A(x))=0$. From the properties $(\textrm{\textbf{Ep}}_\textrm{\textbf{F}}2)$  and $(\textrm{\textbf{Ep}}_\textrm{\textbf{NS}}1)$, we deduce that it should be $h_A(x)|p_A(x)=\{{\frac{1}{2}}|1\}$. 
\\
Proof of axiom ($\textrm{\textbf{Ep}}_\textrm{\textbf{C}}\textrm{3}$): The aim is clear from the properties $(\textrm{\textbf{Ep}}_\textrm{\textbf{F}}3)$  and $(\textrm{\textbf{Ep}}_\textrm{\textbf{NS}}3)$.
\\
Proof of axiom ($\textrm{\textbf{Ep}}_\textrm{\textbf{C}}\textrm{4}$): 
If $h_A^{\sigma(j)}(x)\leq h_B^{\sigma(j)}(x)\leq \frac{1}{2}$ and 
  $\Pi(p_A^{{\sigma(i)}}(x),p_A^{{\sigma(j)}}(x))\leq \Pi(p_B^{{\sigma(i)}}(x),p_B^{{\sigma(j)}}(x))$ 
  or $h_A^{\sigma(j)}(x)\geq h_B^{\sigma(j)}(x)\geq \frac{1}{2}$ and 
  $\Pi(p_A^{{\sigma(i)}}(x),p_A^{{\sigma(j)}}(x))\geq \Pi(p_B^{{\sigma(i)}}(x),p_B^{{\sigma(j)}}(x))$,\\
   together with \\
   if $|h_A^{\sigma(i)}(x)-h_A^{\sigma(j)}(x)|< |h_B^{\sigma(i)}(x)-h_B^{\sigma(j)}(x)|$
  and 
  $\Pi(p_A^{{\sigma(i)}}(x),p_A^{{\sigma(j)}}(x))= \Pi(p_B^{{\sigma(i)}}(x),p_B^{{\sigma(j)}}(x))$; or 
   $|h_A^{\sigma(i)}(x)-h_A^{\sigma(j)}(x)|= |h_B^{\sigma(i)}(x)-h_B^{\sigma(j)}(x)|$
  and 
  $\Pi(p_A^{{\sigma(i)}}(x),p_A^{{\sigma(j)}}(x))> \Pi(p_B^{{\sigma(i)}}(x),p_B^{{\sigma(j)}}(x))$
   for any $i,j=1,2,...,l_x$, then \\
   from the properties $(\textrm{\textbf{Ep}}_\textrm{\textbf{F}}4)$  and $(\textrm{\textbf{Ep}}_\textrm{\textbf{NS}}4)$, we conclude respectively that $Ep_F(h_A(x)|p_A(x)) \leq Ep_F(h_B(x)|p_B(x))$ and $Ep_{NS}(h_A(x)|p_A(x))\leq Ep_{NS}(h_B(x)|p_B(x))$. Consequently, from the property ($\Theta2$) we deduce that
\begin{eqnarray*} 
\Theta(Ep_F(h_A(x)|p_A(x)),Ep_{NS}(h_A(x)|p_A(x)))  \leq \Theta( Ep_F(h_B(x)|p_B(x)), Ep_{NS}(h_B(x)|p_B(x))),
\end{eqnarray*}  
that is, $Ep_C(h_A(x)|p_A(x)) \leq Ep_C(h_B(x)|p_B(x))$. $\Box$

Taking the above theorem into account, we are able to construct different comprehensive entropy measures for PHFEs by using some special mapping $\Theta$ in the form of 
\begin{description}
\item $\Theta_1(x,y)=\max\{x,y\}$;
\item $\Theta_2(x,y)=x+y-xy$;
\item $\Theta_3(x,y)=\min\{x+y,1\}$.
\end{description}
Therefore, we get 
\begin{eqnarray}\label{Ep-C1-1}
&&Ep_{C1}(h_A(x)|p_A(x))=\max\{Ep_F(h_A(x)|p_A(x)),Ep_{NS}(h_A(x)|p_A(x))\};\\
&&Ep_{C2}(h_A(x)|p_A(x))=Ep_F(h_A(x)|p_A(x))+Ep_{NS}(h_A(x)|p_A(x))-Ep_F(h_A(x)|p_A(x))Ep_{NS}(h_A(x)|p_A(x));\nonumber\\&&\\
&&Ep_{C3}(h_A(x)|p_A(x))=\min\{Ep_F(h_A(x)|p_A(x))+Ep_{NS}(h_A(x)|p_A(x)),1\}.\label{Ep-C1-3}
\end{eqnarray}

\subsection{{Entropy-based distance measure for PHFEs}}

In this part of the contribution, we will introduce some entropy-based distance measures for PHFEs in the same line of distance measures
for hesitant fuzzy information proposed in \cite{[5EntM]}. Before going more in details with the problem, let us present the axiomatic definition of distance
measure for PHFEs.
\begin{Definition}\label{pro-dp}
Suppose that $h_A(x)|p_A(x)=\{h^{\sigma(i)}_A(x)|p^{\sigma(i)}_A(x)\}_{i=1}^{l_{xA}}$ and $h_B(x)|p_B(x)=\{h^{\sigma(i)}_B(x)|p^{\sigma(i)}_B(x)\}_{i=1}^{l_{xB}}$ are two PHFEs on $X$. Then, the mapping $Dp$ is called a distance measure for PHFEs  
if it possesses the following properties:
\begin{description}
\item[($\textrm{\textbf{Dp}}$0)] $0\leq Dp(h_A(x)|p_A(x), h_B(x)|p_B(x))\leq 1$;
  \item[($\textrm{\textbf{Dp}}$1)] $Dp(h_A(x)|p_A(x), h_B(x)|p_B(x))=0$ if and only if $h_A(x)|p_A(x)=h_B(x)|p_B(x)$;
  \item[($\textrm{\textbf{Dp}}$2)] $Dp(h_A(x)|p_A(x), h_B(x)|p_B(x))=Dp(h_B(x)|p_B(x), h_A(x)|p_A(x))$.
\end{description}
\end{Definition}

Now, in order to make a connection between PHFE distance and PHFE entropy measures, we are required to introduce the hybrid form of PHFEs $h_A(x)|p_A(x)=\{h^{\sigma(i)}_A(x)|p^{\sigma(i)}_A(x)\}_{i=1}^{l_{xA}}$ and $h_B(x)|p_B(x)=\{h^{\sigma(i)}_B(x)|p^{\sigma(i)}_B(x)\}_{i=1}^{l_{xB}}$ as the following:\begin{eqnarray}\label{d-1}
h^{\sigma(k)}_{\varrho(A,B)}(x)|p^{\sigma(k)}_{\varrho(A,B)}(x)=
\frac{1-|h_A^{\sigma(i)}(x)-h_B^{\sigma(j)}(x)|}{2}|\Pi(p_A^{{\sigma(i)}}(x),p_B^{{\sigma(j)}}(x)), \quad k=1,2,...,l_{xA}\times l_{xB},
\end{eqnarray}
and therefore, the hybrid form of PHFEs $h_A(x)|p_A(x)$ and $h_B(x)|p_B(x)$ will be as the form of $h_{\varrho(A,B)}(x)|p_{\varrho(A,B)}(x)=\{h^{\sigma(k)}_{\varrho(A,B)}(x)|p^{\sigma(k)}_{\varrho(A,B)}(x)\}$.
\\
It can be easily observed that
\begin{eqnarray*} 
h_{\varrho(A^c,B^c)}(x)|p_{\varrho(A^c,B^c)}(x)=h_{\varrho(A,B)}(x)|p_{\varrho(A,B)}(x). 
\end{eqnarray*} 

\begin{Theorem}
Suppose that $h_A(x)|p_A(x)=\{h^{\sigma(i)}_A(x)|p^{\sigma(i)}_A(x)\}_{i=1}^{l_{xA}}$ and $h_B(x)|p_B(x)=\{h^{\sigma(i)}_B(x)|p^{\sigma(i)}_B(x)\}_{i=1}^{l_{xB}}$ are two PHFEs on $X$. 
Moreover, we assume that $\psi:[0,1]\rightarrow [0,1]$ is a strictly monotone increasing real function, and $Ep_C$ is an entropy measure between PHFEs.
Then, the mapping   
\begin{eqnarray}\label{def-dp} 
Dp(h_A(x)|p_A(x),h_B(x)|p_B(x))=1-\frac{\psi(Ep_C(h_{\varrho(A,B)}(x)|p_{\varrho(A,B)}(x))-\psi(0)}{\psi(1)-\psi(0)}
\end{eqnarray}
is a distance measure for PHFEs.
\end{Theorem}
\textit{Proof.} We need to show that $Dp(h_A(x)|p_A(x),h_B(x)|p_B(x))$ satisfies the requirements ($\textrm{\textbf{Dp}}$0)-($\textrm{\textbf{Dp}}$2) listed in Definition \ref{pro-dp}.
\\
Proof of axiom ($\textrm{\textbf{Dp}}$0): It is straightforward. 
\\
Proof of axiom ($\textrm{\textbf{Dp}}$1): By taking the definition of $Dp(h_A(x)|p_A(x),h_B(x)|p_B(x))$ given by (\ref{def-dp}) into account, we deduce that
\begin{eqnarray*}
Dp(h_A(x)|p_A(x),h_B(x)|p_B(x))=1-\frac{\psi(Ep(h_{\varrho(A,B)}(x)|p_{\varrho(A,B)}(x))-\psi(0)}{\psi(1)-\psi(0)}=0
\end{eqnarray*}
if and only if 
\begin{eqnarray*}
\frac{\psi(Ep_C(h_{\varrho(A,B)}(x)|p_{\varrho(A,B)}(x))-\psi(0)}{\psi(1)-\psi(0)}=1
\end{eqnarray*}
if and only if 
\begin{eqnarray*}
\psi(Ep_C(h_{\varrho(A,B)}(x)|p_{\varrho(A,B)}(x))=\psi(1)
\end{eqnarray*}
if and only if 
\begin{eqnarray*}
Ep_C(h_{\varrho(A,B)}(x)|p_{\varrho(A,B)}(x)=1.
\end{eqnarray*}
Now, from the property ($\textrm{\textbf{Ep}}_\textrm{\textbf{C}}$2) of Definition \ref{comp-ent}, we conclude that 
$Ep_C(h_{\varrho(A,B)}(x)|p_{\varrho(A,B)}(x))=1$ if and only if 
\begin{eqnarray*}
&&h_{\varrho(A,B)}(x)|p_{\varrho(A,B)}(x)=\{\frac{1}{2}|1\}.
\end{eqnarray*}
This is held if and only if 
\begin{eqnarray*}
&&\frac{1-|h_A^{\sigma(i)}(x)-h_B^{\sigma(j)}(x)|}{2}=\frac{1}{2}, \quad i=1,2,...,l_{xA},j=1,2,..., l_{xB}.
\end{eqnarray*}
which implies that $h_A(x)|p_A(x)=h_B(x)|p_B(x)$.
\\
Proof of axiom ($\textrm{\textbf{Dp}}$3): It is obvious. $\Box$ 

If we take different forms of strictly monotone increasing function $\psi$ as $\psi(\zeta)=\zeta$; $\psi(\zeta)=\zeta^2$; $\psi(\zeta)=\frac{2\zeta}{1+\zeta}$; or $\psi(\zeta)=\zeta e^{\zeta-1}$ together with different kinds of $Ep_C$ given by (\ref{Ep-C1-1})-(\ref{Ep-C1-3}), we then are able to construct a variety of distance measures for PHFEs.


\section{\textcolor{blue}{A comparative analysis of PHFE entropy measures}}\label{sec4}

The current section intends to present a comparative analysis between the performance of well-established entropy measures for PHFEs and some other existing PHFE entropy measures.
\\
Although a lot of works \cite{ding,liu,pang,[anov],su,[31yue]} have dealt with PHFS in describing the practical
condition, but there are not enough contributions available in which the PHFS entropy measure for dealing with uncertainty of
information has been investigated so far.
To the author's best knowledge, Su et al.'s \cite{su} work is one of the rare report that aims
 at introducing two classes of entropy measure for PHFEs including (i) the two membership degree-based entropy measures \textcolor{blue}{$E_{P_1}$ and $E_{P_2}$}, and (ii) the distance-based entropy measure \textcolor{blue}{$E_{D}$} 
\textcolor{blue}{
which are respectively given as follows:
\begin{eqnarray}\label{Ep-1}
&&E_{P_1}(h(x)|p(x))=-\frac{1}{Ln(2)}\sum_{i=1}^{l_x} [h^{\sigma(i)}(x) Ln(h^{\sigma(i)}(x))+(1-h^{\sigma(i)}(x)) Ln(1-h^{\sigma(i)}(x))]p^{{\sigma(i)}}(x);\\
&&E_{P_2}(h(x)|p(x))=\frac{1}{(\sqrt{e}-1)}\sum_{i=1}^{l_x} [h^{\sigma(i)}(x) e^{1-h^{\sigma(i)}(x)}+(1-h^{\sigma(i)}(x)) e^{h^{\sigma(i)}(x)}-1]
p^{{\sigma(i)}}(x);\label{Ep-2}\\
&&E_D(h(x)|p(x))=\zeta(d_{SU}(h(x)|p(x),\frac{1}{2}|1)),\label{Ep-d}
\end{eqnarray}
where $\zeta$ stands for a strictly monotone decreasing function with the conditions $\zeta(0)=1$ and $\zeta(\frac{1}{2})=0$.
Furthermore, the probabilistic hesitant like-distance measure $d_{SU}$ between the PHFEs $h_A(x)|p_A(x)=\{h^{\sigma(i)}_A(x)|p^{\sigma(i)}_A(x)\}_{i=1}^{l_{xA}}$ and $h_B(x)|p_B(x)=\{h^{\sigma(i)}_B(x)|p^{\sigma(i)}_B(x)\}_{i=1}^{l_{xB}}$ is defined in terms of 
\begin{eqnarray}\label{dsu}
&&d_{SU}(h_A(x)|p_A(x),h_B(x)|p_B(x))=|\sum_{i=1}^{l_{xA}}h^{\sigma(i)}_A(x)p^{\sigma(i)}_A(x)-\sum_{i=1}^{l_{xB}}h^{\sigma(i)}_B(x)p^{\sigma(i)}_B(x)|.
\end{eqnarray}
}\\
By the way, each of \textcolor{blue}{the above-mentioned entropy measures} has the relevance \textcolor{blue}{shortcomings which are} discussed here as the \textcolor{blue}{followings.}
\\
Part (i): According to the fourth property $(EP4)$ of Definition 4 in \cite{su}, the membership degree-based entropy measures do not explain about the case where the corresponding probabilities are different. More specifically, the first part of the property $(EP4)$ is based on the equality relationship of the corresponding probabilities, and this condition is abandoned in the second part. Furthermore, Su et al. 
admitted that  the membership degree-based entropy measures \textcolor{blue}{$E_{P_1}$ and $E_{P_2}$} are invalid in the case where $h(x)|p(x)=\{0|0.5,1|0.5\}$. 
\\
Part (ii): Unlike the traditional distance measures, Su et al. \cite{su} employed the concept of like-distance measure in constructing the distance-based entropy measure for PHFEs. 
{In the special case} where PHFEs are reduced to HFEs, that is, $p^{\sigma(i)}_A(x)=\frac{1}{l_{xA}}$ (for any $i=1,2,...,{l_{xA}}$) and $p^{\sigma(i)}_B(x)=\frac{1}{l_{xB}}$ (for any $i=1,2,...,{l_{xB}}$), the above distance measure \textcolor{blue}{$d_{SU}$} should be converted to a distance measure for HFEs. Meanwhile, by the latter assumption, the distance \textcolor{blue}{$d_{SU}$} is \textcolor{blue}{reduced} to  
\begin{eqnarray*}
&&d_{SU}(h_A(x),h_B(x))=|\frac{1}{{l_{xA}}}\sum_{i=1}^{l_{xA}}h^{\sigma(i)}_A(x)-\frac{1}{{l_{xB}}}\sum_{i=1}^{l_{xB}}h^{\sigma(i)}_B(x)|
\end{eqnarray*}
which does not coincide with the traditional definition of distance measure {(see e.g. \cite{[5EntM]})}
\begin{eqnarray*}
&&d(h_A(x),h_B(x))=\frac{1}{{l_{xA}}}\frac{1}{{l_{xB}}}\sum_{i=1}^{{l_{xA}}}\sum_{j=1}^{{l_{xB}}}|h^{\sigma(i)}_A(x)-h^{\sigma(j)}_B(x)|.
\end{eqnarray*}

In this portion of the contribution, we are going to re-consider the results of Su et al. \cite{su}, and compare them with that of the proposed entropy measures for PHFEs.


\begin{Example}\label{ex1} 
{\rm{Re-consider the following PHFEs \textcolor{blue}{which were} given initially \textcolor{blue}{in \cite{su}}:}} 

\begin{eqnarray*}
&&h_1(x)|p_1(x)=\{0.7|0.2,0.9|0.8\},\\
&&h_2(x)|p_2(x)=\{0.6|0.9,0.9|0.1\},\\
&&h_3(x)|p_3(x)=\{0.6|0.1,0.9|0.9\}.
\end{eqnarray*}
\end{Example} 
As explained and demonstrated below, the proposed fuzziness-based entropy measures $Ep_{1F}$ and $Ep_{2F}$ can produce better results in distinguishing the above PHFEs
than Su et al's \cite{su} entropy measures.
From Figure 1, we intuitively observe that  $h_2(x)|p_2(x)$ and  $h_3(x)|p_3(x)$ inherit more fuzziness 
than $h_1(x)|p_1(x)$, that is, the value of HFEs $h_2(x)$ and $h_3(x)$ are larger than the value of HFE $h_1(x)$. Moreover, 
in comparing $h_2(x)|p_2(x)$ and  $h_3(x)|p_3(x)$, it is obvious that the probabilities from the left- to right-side of origin (blue to red circles) are in lexicographical-decreasing order. Therefore, $h_2(x)|p_2(x)$ is {greater} from fuzziness degree rather than $h_3(x)|p_3(x)$.
\\
Thus, it is clearly seen that no matter which fuzzy entropy
measure is employed, the fuzzy entropy measure of $h_2(x)|p_2(x)$ should be greater than that of $h_3(x)|p_3(x)$, and both fuzzy entropy measures of them should be greater than that of $h_1(x)|p_1(x)$. This fact is observable in Table 1, meanwhile, Su et al's \cite{su} results are not in accordance with this fact.

\newpage
\begin{center}
{Table 1. Fuzziness-based entropy measures for $h_i(x)|p_i(x)$.}\\
{\small{
\begin{tabular}{|c| c c c| c|}
 \hline
 Entropy &  $h_1(x)|p_1(x)$  &  $h_2(x)|p_2(x)$  & $h_3(x)|p_3(x)$  &Ranking ~order\\ \hline
 \hline 
 $E_{P1}$  \cite{su} & 0.551 &   0.921 &   0.519& $E_{P1}(h_2(x)|p_2(x))>E_{P1}(h_1(x)|p_1(x))>E_{P1}(h_3(x)|p_3(x))$\\ \hline
 $E_{P2}$   \cite{su} & 0.466&    0.903 &   0.430& $E_{P2}(h_2(x)|p_2(x))>E_{P2}(h_1(x)|p_1(x))>E_{P2}(h_3(x)|p_3(x))$\\ \hline
 \hline  
$Ep_{1F}$  & 0.1513 &   0.3488 &   0.1945& $Ep_{1F}(h_2(x)|p_2(x))>Ep_{1F}(h_3(x)|p_3(x))>Ep_{1F}(h_1(x)|p_1(x))$\\ \hline
$Ep_{2F}$                 & 0.4061&    0.7177 &   0.4922& $Ep_{2F}(h_2(x)|p_2(x))>Ep_{2F}(h_3(x)|p_3(x))>Ep_{2F}(h_1(x)|p_1(x))$\\ \hline

\end{tabular}
}}\vspace{0.1cm}
\end{center}

\begin{figure}
  \includegraphics{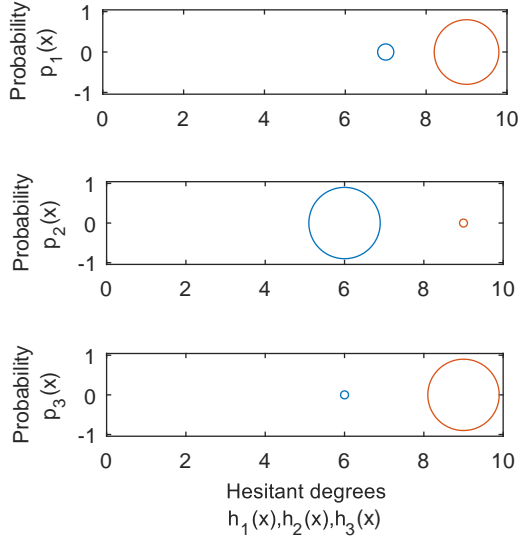}\\
   \vspace{-1. cm}
 \caption{The schematic of $h_i(x)|p_i(x)$ (for $i=1,2,3$).}\label{fig1}
\end{figure}

\begin{Example}\label{ex2} 
\rm{We consider the following PHFEs:}
\begin{eqnarray*}
&&h_1(x)|p_1(x)=\{0.4|0.5,0.6|0.5\},\\
&&h_2(x)|p_2(x)=\{0.2|0.5,0.8|0.5\}.
\end{eqnarray*}
\end{Example} 
It is interesting to note that the proposed non-specific-based entropy measures $Ep_{1NS}$, $Ep_{2NS}$  and $Ep_{3NS}$ produce better results in distinguishing the above PHFEs
than Su et al's \cite{su} entropy measures. This is obvious from Figure 2 where we intuitively observe that  $h_2(x)|p_2(x)$ inherits more non-specificity than $h_1(x)|p_1(x)$,
that is, the distance between elements involved in $h_2(x)$ is larger than those in $h_1(x)$.
\\
Thus, it is clearly seen that no matter which non-specific entropy
measure is employed, the  entropy measure of $h_2(x)|p_2(x)$ should be greater than that of $h_1(x)|p_1(x)$. This fact is observable in Table 2, meanwhile, Su et al's \cite{su} results are not in accordance with this fact.

\newpage
\begin{center}
{Table 2. Non-specific-based entropy measures for $h_i(x)|p_i(x)$.}\\
{\small{
\begin{tabular}{|c| c c | c|}
 \hline
 Entropy &  $h_1(x)|p_1(x)$  &  $h_2(x)|p_2(x)$  &Ranking ~order\\ \hline
 \hline 
 $E_{D}$   \cite{su} & 1&   1 &    $E_{D}(h_1(x)|p_1(x))=E_{D}(h_2(x)|p_2(x))$\\ \hline
 \hline  
$Ep_{1NS}$  & 0.1710 &   0.2745 &   $Ep_{1NS}(h_1(x)|p_1(x))<Ep_{1NS}(h_2(x)|p_2(x))$\\ \hline
$Ep_{2NS}$                 & 0.0999 &   0.2114 &   $Ep_{2NS}(h_1(x)|p_1(x))<Ep_{2NS}(h_2(x)|p_2(x))$\\ \hline
$Ep_{3NS}$   & 0.1710 &   0.2750 &   $Ep_{3NS}(h_1(x)|p_1(x))<Ep_{3NS}(h_2(x)|p_2(x))$\\ \hline
\end{tabular}\\
}}\vspace{0.1cm}
\end{center}

\begin{figure}
 \includegraphics{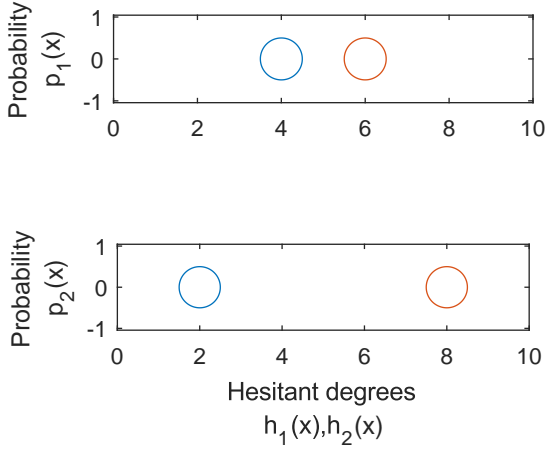}\\
 \vspace{-4.5 cm}
 \caption{The schematic of $h_i(x)|p_i(x)$ (for $i=1,2$).}\label{fig2}
\end{figure}

In what follows, we employ both sets of PHFEs presented in Example \ref{ex1} and  Example \ref{ex2} to explain and demonstrate the ability of 
comprehensive entropy measures which are constructed based on both fuzziness and non-specificity aspects.
\begin{Example} \rm{Let}
\begin{eqnarray*}
&&h_1(x)|p_1(x)=\{0.7|0.2,0.9|0.8\},\\
&&h_2(x)|p_2(x)=\{0.6|0.9,0.9|0.1\},\\
&&h_3(x)|p_3(x)=\{0.6|0.1,0.9|0.9\},\\
&&h_4(x)|p_4(x)=\{0.4|0.5,0.6|0.5\},\\
&&h_5(x)|p_5(x)=\{0.2|0.5,0.8|0.5\},
\end{eqnarray*}
\end{Example} 
be five PHFEs. If we implement the comprehensive entropy measures $Ep_{C1}$, $Ep_{C2}$ and $Ep_{C3}$ based on composition of different $Ep_{F}$s and $Ep_{NS}$s, then the results will be as those given in Tables 3,4,...,8.
\\
\textit{Case 1.} Suppose that 
\begin{eqnarray*}
&&Ep_{C11}(h_A(x)|p_A(x))=\max\{Ep_{1F}(h_A(x)|p_A(x)),Ep_{1NS}(h_A(x)|p_A(x))\},\\
&&Ep_{C12}(h_A(x)|p_A(x))=\max\{Ep_{1F}(h_A(x)|p_A(x)),Ep_{2NS}(h_A(x)|p_A(x))\},\\
&&Ep_{C13}(h_A(x)|p_A(x))=\max\{Ep_{1F}(h_A(x)|p_A(x)),Ep_{3NS}(h_A(x)|p_A(x))\}.
\end{eqnarray*}
Then, the results are that given in Table 3.

\begin{center}
{Table 3. Combination of fuzziness and Non-specificity entropies for $h_i(x)|p_i(x)$.}\\
{\small{
\begin{tabular}{|c| c c c c c |}
 \hline
 Entropy &  $h_1(x)|p_1(x)$  &  $h_2(x)|p_2(x)$ & $h_3(x)|p_3(x)$  &  $h_4(x)|p_4(x)$ & $h_5(x)|p_5(x)$ \\
 & &&Rankings &&\\ \hline
 \hline 
 $E_{D}$  \cite{su} & * &   *&   *&   * &   *  \\
 &$E_{D}(h_4(x)|p_4(x))$&$>E_{D}(h_5(x)|p_5(x))$&$>E_{D}(h_2(x)|p_2(x))$&$>E_{D}(h_1(x)|p_1(x))$&$>E_{D}(h_3(x)|p_3(x))$\\ \hline
 \hline  
$Ep_{C11}$  & 0.2676 &   0.3488 &   0.1945&    0.4126 &   0.2887 \\
&$Ep_{C11}(h_4(x)|p_4(x))$&$>Ep_{C11}(h_2(x)|p_2(x))$&$>Ep_{C11}(h_5(x)|p_5(x))$&$>Ep_{C11}(h_1(x)|p_1(x))$&$>Ep_{C11}(h_3(x)|p_3(x))$\\ \hline
$Ep_{C12}$   & 0.2552&    0.3488&    0.1945&    0.4126 &   0.2745 \\
&$Ep_{C12}(h_4(x)|p_4(x))$&$>Ep_{C12}(h_2(x)|p_2(x))$&$>Ep_{C12}(h_5(x)|p_5(x))$&$>Ep_{C12}(h_1(x)|p_1(x))$&$>Ep_{C12}(h_3(x)|p_3(x))$\\ 
\hline
$Ep_{C13}$  & 0.2059&    0.3488 &   0.1945&    0.4126&    0.2443\\
& $Ep_{C13}(h_4(x)|p_4(x))$&$>Ep_{C13}(h_2(x)|p_2(x))$&$>Ep_{C13}(h_5(x)|p_5(x))$&$>Ep_{C13}(h_1(x)|p_1(x))$&$>Ep_{C13}(h_3(x)|p_3(x))$
 \\ \hline
\end{tabular}\\
"*" means cannot distinguish directly.
}}\vspace{0.1cm}
\end{center}

\textit{Case 2.} Suppose that 
\begin{eqnarray*}
&&Ep_{C21}(h_A(x)|p_A(x))=\max\{Ep_{2F}(h_A(x)|p_A(x)),Ep_{1NS}(h_A(x)|p_A(x))\},\\
&&Ep_{C22}(h_A(x)|p_A(x))=\max\{Ep_{2F}(h_A(x)|p_A(x)),Ep_{2NS}(h_A(x)|p_A(x))\},\\
&&Ep_{C23}(h_A(x)|p_A(x))=\max\{Ep_{2F}(h_A(x)|p_A(x)),Ep_{3NS}(h_A(x)|p_A(x))\}.
\end{eqnarray*}
Then, the results are that given in Table 4.    

%
%

\begin{center}
{Table 4. Combination of fuzziness and Non-specificity entropies for $h_i(x)|p_i(x)$.}\\
{\small{
\begin{tabular}{|c| c c c c c |}
 \hline
 Entropy &  $h_1(x)|p_1(x)$  &  $h_2(x)|p_2(x)$ & $h_3(x)|p_3(x)$  &  $h_4(x)|p_4(x)$ & $h_5(x)|p_5(x)$ \\
 & &&Rankings &&\\ \hline
 \hline 
 $E_{D}$  \cite{su} & * &   *&   *&   * &   *  \\
 &$E_{D}(h_4(x)|p_4(x))$&$>E_{D}(h_5(x)|p_5(x))$&$>E_{D}(h_2(x)|p_2(x))$&$>E_{D}(h_1(x)|p_1(x))$&$>E_{D}(h_3(x)|p_3(x))$\\ \hline
 \hline  
$Ep_{C21}$  & 0.4061&    0.7177 &   0.4922 &   0.7140 &   0.5057 \\
&$Ep_{C21}(h_2(x)|p_2(x))$&$>Ep_{C21}(h_4(x)|p_4(x))$&$>Ep_{C21}(h_5(x)|p_5(x))$&$>Ep_{C21}(h_3(x)|p_3(x))$&$>Ep_{C21}(h_1(x)|p_1(x))$\\ \hline
$Ep_{C22}$   &  0.4061&    0.7177 &   0.4922&    0.7140&    0.5057 \\
&$Ep_{C22}(h_2(x)|p_2(x))$&$>Ep_{C22}(h_4(x)|p_4(x))$&$>Ep_{C22}(h_5(x)|p_5(x))$&$>Ep_{C22}(h_3(x)|p_3(x))$&$>Ep_{C22}(h_1(x)|p_1(x))$\\ 
\hline
$Ep_{C23}$  & 0.4061&    0.7177&    0.4922 &   0.7140&    0.5057\\
&$Ep_{C23}(h_2(x)|p_2(x))$&$>Ep_{C23}(h_4(x)|p_4(x))$&$>Ep_{C23}(h_5(x)|p_5(x))$&$>Ep_{C23}(h_3(x)|p_3(x))$&$>Ep_{C23}(h_1(x)|p_1(x))$\\
\hline
\end{tabular}\\
"*" means cannot distinguish directly.
}}\vspace{0.1cm}
\end{center}

\textit{Case 3.} Suppose that 
\begin{eqnarray*}
&&Ep_{C11}(h_A(x)|p_A(x))=Ep_{1F}(h_A(x)|p_A(x))+Ep_{1NS}(h_A(x)|p_A(x))-Ep_{1F}(h_A(x)|p_A(x))Ep_{1NS}(h_A(x)|p_A(x)),\\
&&Ep_{C12}(h_A(x)|p_A(x))=Ep_{1F}(h_A(x)|p_A(x))+Ep_{2NS}(h_A(x)|p_A(x))-Ep_{1F}(h_A(x)|p_A(x))Ep_{2NS}(h_A(x)|p_A(x)),\\
&&Ep_{C13}(h_A(x)|p_A(x))=Ep_{1F}(h_A(x)|p_A(x))+Ep_{3NS}(h_A(x)|p_A(x))-Ep_{1F}(h_A(x)|p_A(x))Ep_{3NS}(h_A(x)|p_A(x)).
\end{eqnarray*}
Then, the results are that given in Table 5.   

%

\begin{center}
{Table 5. Combination of fuzziness and Non-specificity entropies for $h_i(x)|p_i(x)$.}\\
{\small{
\begin{tabular}{|c| c c c c c |}
 \hline
 Entropy &  $h_1(x)|p_1(x)$  &  $h_2(x)|p_2(x)$ & $h_3(x)|p_3(x)$  &  $h_4(x)|p_4(x)$ & $h_5(x)|p_5(x)$ \\
 & &&Rankings &&\\ \hline
 \hline 
 $E_{D}$  \cite{su} & * &   *&   *&   * &   *  \\
 &$E_{D}(h_4(x)|p_4(x))$&$>E_{D}(h_5(x)|p_5(x))$&$>E_{D}(h_2(x)|p_2(x))$&$>E_{D}(h_1(x)|p_1(x))$&$>E_{D}(h_3(x)|p_3(x))$\\ \hline
 \hline  
$Ep_{C11}$  & 0.3784&    0.4570&    0.3391&    0.5256&    0.4624 \\
&$Ep_{C11}(h_4(x)|p_4(x))$&$>Ep_{C11}(h_5(x)|p_5(x))$&$>Ep_{C11}(h_2(x)|p_2(x))$&$>Ep_{C11}(h_1(x)|p_1(x))$&$>Ep_{C11}(h_3(x)|p_3(x))$\\ \hline
$Ep_{C12}$   &  0.3679&    0.4393&    0.3179&    0.5130 &   0.4517 \\
&$Ep_{C12}(h_4(x)|p_4(x))$&$>Ep_{C12}(h_5(x)|p_5(x))$&$>Ep_{C12}(h_2(x)|p_2(x))$&$>Ep_{C12}(h_1(x)|p_1(x))$&$>Ep_{C12}(h_3(x)|p_3(x))$\\ 
\hline
$Ep_{C13}$  & 0.3260&    0.3879&   0.2530&    0.4713&    0.4040\\
&$Ep_{C13}(h_4(x)|p_4(x))$&$>Ep_{C13}(h_5(x)|p_5(x))$&$>Ep_{C13}(h_2(x)|p_2(x))$&$>Ep_{C13}(h_1(x)|p_1(x))$&$>Ep_{C13}(h_3(x)|p_3(x))$\\
\hline
\end{tabular}\\
"*" means cannot distinguish directly.
}}\vspace{0.1cm}
\end{center}

\textit{Case 4.} Suppose that 
\begin{eqnarray*}
&&Ep_{C21}(h_A(x)|p_A(x))=Ep_{2F}(h_A(x)|p_A(x))+Ep_{1NS}(h_A(x)|p_A(x))-Ep_{2F}(h_A(x)|p_A(x))Ep_{1NS}(h_A(x)|p_A(x)),\\
&&Ep_{C22}(h_A(x)|p_A(x))=Ep_{2F}(h_A(x)|p_A(x))+Ep_{2NS}(h_A(x)|p_A(x))-Ep_{2F}(h_A(x)|p_A(x))Ep_{2NS}(h_A(x)|p_A(x)),\\
&&Ep_{C23}(h_A(x)|p_A(x))=Ep_{2F}(h_A(x)|p_A(x))+Ep_{3NS}(h_A(x)|p_A(x))-Ep_{2F}(h_A(x)|p_A(x))Ep_{3NS}(h_A(x)|p_A(x)).
\end{eqnarray*}
Then, the results are that given in Table 6.

%

\begin{center}
{Table 6. Combination of fuzziness and Non-specificity entropies for $h_i(x)|p_i(x)$.}\\
{\small{
\begin{tabular}{|c| c c c c c |}
 \hline
 Entropy &  $h_1(x)|p_1(x)$  &  $h_2(x)|p_2(x)$ & $h_3(x)|p_3(x)$  &  $h_4(x)|p_4(x)$ & $h_5(x)|p_5(x)$ \\
 & &&Rankings &&\\ \hline
 \hline 
 $E_{D}$  \cite{su} & * &   *&   *&   * &   *  \\
 &$E_{D}(h_4(x)|p_4(x))$&$>E_{D}(h_5(x)|p_5(x))$&$>E_{D}(h_2(x)|p_2(x))$&$>E_{D}(h_1(x)|p_1(x))$&$>E_{D}(h_3(x)|p_3(x))$\\ \hline
 \hline  
$Ep_{C21}$  & 0.5650&    0.7646&    0.5834 &   0.7691&    0.6484 \\
&$Ep_{C21}(h_4(x)|p_4(x))$&$>Ep_{C21}(h_2(x)|p_2(x))$&$>Ep_{C21}(h_5(x)|p_5(x))$&$>Ep_{C21}(h_3(x)|p_3(x))$&$>Ep_{C21}(h_1(x)|p_1(x))$\\ \hline
$Ep_{C22}$   &  0.5577&    0.7569&    0.5701&   0.7629 &   0.6414 \\
&$Ep_{C22}(h_4(x)|p_4(x))$&$>Ep_{C22}(h_2(x)|p_2(x))$&$>Ep_{C22}(h_5(x)|p_5(x))$&$>Ep_{C22}(h_3(x)|p_3(x))$&$>Ep_{C22}(h_1(x)|p_1(x))$\\ 
\hline
$Ep_{C23}$  & 0.5284&    0.7346&    0.5291 &   0.7426&    0.6102\\
&$Ep_{C23}(h_4(x)|p_4(x))$&$>Ep_{C23}(h_2(x)|p_2(x))$&$>Ep_{C23}(h_5(x)|p_5(x))$&$>Ep_{C23}(h_3(x)|p_3(x))$&$>Ep_{C23}(h_1(x)|p_1(x))$\\
\hline
\end{tabular}\\
"*" means cannot distinguish directly.
}}\vspace{0.1cm}
\end{center}

\textit{Case 5.} Suppose that 
\begin{eqnarray*}
&&Ep_{C11}(h_A(x)|p_A(x))=\min\{Ep_{1F}(h_A(x)|p_A(x))+Ep_{1NS}(h_A(x)|p_A(x)),1 \},\\
&&Ep_{C12}(h_A(x)|p_A(x))=\min\{Ep_{1F}(h_A(x)|p_A(x))+Ep_{2NS}(h_A(x)|p_A(x)),1\},\\
&&Ep_{C13}(h_A(x)|p_A(x))=\min\{Ep_{1F}(h_A(x)|p_A(x))+Ep_{3NS}(h_A(x)|p_A(x)),1\}.
\end{eqnarray*}
Then, the results are that given in Table 7.

%

\begin{center}
{Table 7. Combination of fuzziness and Non-specificity entropies for $h_i(x)|p_i(x)$.}\\
{\small{
\begin{tabular}{|c| c c c c c |}
 \hline
 Entropy &  $h_1(x)|p_1(x)$  &  $h_2(x)|p_2(x)$ & $h_3(x)|p_3(x)$  &  $h_4(x)|p_4(x)$ & $h_5(x)|p_5(x)$ \\
 & &&Rankings &&\\ \hline
 \hline 
 $E_{D}$  \cite{su} & * &   *&   *&   * &   *  \\
 &$E_{D}(h_4(x)|p_4(x))$&$>E_{D}(h_5(x)|p_5(x))$&$>E_{D}(h_2(x)|p_2(x))$&$>E_{D}(h_1(x)|p_1(x))$&$>E_{D}(h_3(x)|p_3(x))$\\ \hline
 \hline  
$Ep_{C11}$  & 0.4189&    0.5150&    0.3740&    0.6050&    0.5329  \\
&$Ep_{C11}(h_4(x)|p_4(x))$&$>Ep_{C11}(h_5(x)|p_5(x))$&$>Ep_{C11}(h_2(x)|p_2(x))$&$>Ep_{C11}(h_1(x)|p_1(x))$&$>Ep_{C11}(h_3(x)|p_3(x))$\\ \hline
$Ep_{C12}$   &  0.4065&    0.4878&    0.3477&    0.5835 &   0.5188 \\
&$Ep_{C12}(h_4(x)|p_4(x))$&$>Ep_{C12}(h_5(x)|p_5(x))$&$>Ep_{C12}(h_2(x)|p_2(x))$&$>Ep_{C12}(h_1(x)|p_1(x))$&$>Ep_{C12}(h_3(x)|p_3(x))$\\
\hline
$Ep_{C13}$  & 0.3572&    0.4089&    0.2671&    0.5125&    0.4557\\
&$Ep_{C13}(h_4(x)|p_4(x))$&$>Ep_{C13}(h_5(x)|p_5(x))$&$>Ep_{C13}(h_2(x)|p_2(x))$&$>Ep_{C13}(h_1(x)|p_1(x))$&$>Ep_{C13}(h_3(x)|p_3(x))$\\
\hline
\end{tabular}\\
"*" means cannot distinguish directly.
}}\vspace{0.1cm}
\end{center}

\textit{Case 6.} Suppose that 
\begin{eqnarray*}
&&Ep_{C21}(h_A(x)|p_A(x))=\min\{Ep_{2F}(h_A(x)|p_A(x))+Ep_{1NS}(h_A(x)|p_A(x)),1\},\\
&&Ep_{C22}(h_A(x)|p_A(x))=\min\{Ep_{2F}(h_A(x)|p_A(x))+Ep_{2NS}(h_A(x)|p_A(x)),1\},\\
&&Ep_{C23}(h_A(x)|p_A(x))=\min\{Ep_{2F}(h_A(x)|p_A(x))+Ep_{3NS}(h_A(x)|p_A(x)),1\}.
\end{eqnarray*}
Then, the results are that given in Table 8.

%
%
\begin{center}
{Table 8. Combination of fuzziness and Non-specificity entropies for $h_i(x)|p_i(x)$.}\\
{\small{
\begin{tabular}{|c| c c c c c |}
 \hline
 Entropy &  $h_1(x)|p_1(x)$  &  $h_2(x)|p_2(x)$ & $h_3(x)|p_3(x)$  &  $h_4(x)|p_4(x)$ & $h_5(x)|p_5(x)$ \\
 & &&Rankings &&\\ \hline
 \hline 
 $E_{D}$  \cite{su} & * &   *&   *&   * &   *  \\
 &$E_{D}(h_4(x)|p_4(x))$&$>E_{D}(h_5(x)|p_5(x))$&$>E_{D}(h_2(x)|p_2(x))$&$>E_{D}(h_1(x)|p_1(x))$&$>E_{D}(h_3(x)|p_3(x))$\\ \hline
 \hline  
$Ep_{C21}$  & 0.6737&    0.8839&    0.6718&    0.9065&    0.7944  \\
&$Ep_{C21}(h_4(x)|p_4(x))$&$>Ep_{C21}(h_2(x)|p_2(x))$&$>Ep_{C21}(h_5(x)|p_5(x))$&$>Ep_{C21}(h_1(x)|p_1(x))$&$>Ep_{C21}(h_3(x)|p_3(x))$\\ \hline
$Ep_{C22}$   & 0.6613&    0.8567&    0.6455&    0.8850&    0.7802\\
&$Ep_{C22}(h_4(x)|p_4(x))$&$>Ep_{C22}(h_2(x)|p_2(x))$&$>Ep_{C22}(h_5(x)|p_5(x))$&$>Ep_{C22}(h_1(x)|p_1(x))$&$>Ep_{C22}(h_3(x)|p_3(x))$\\
\hline
$Ep_{C23}$  & 0.6120&    0.7777&    0.5649&    0.8140&    0.7171\\
&$Ep_{C23}(h_4(x)|p_4(x))$&$>Ep_{C23}(h_2(x)|p_2(x))$&$>Ep_{C23}(h_5(x)|p_5(x))$&$>Ep_{C23}(h_1(x)|p_1(x))$&$>Ep_{C23}(h_3(x)|p_3(x))$\\
\hline
\end{tabular}\\
"*" means cannot distinguish directly.
}}\vspace{0.1cm}
\end{center}

From Tables 3,4,...,8, one observes that the entropy measure $E_{D}$  \cite{su} cannot distinguish directly (non-algorithmically) between the worst and the best alternatives while the proposed entropy measures can do this effectively.

\section{\textcolor{blue}{Multiple criteria decision-making with PHFE information}}\label{sec5}

For providing further insights about the proposed entropy measures of PHFEs, we compare them against the other existing entropy measures for PHFEs in form of a multiple criteria decision making (MCDM) problem.
Generally, a MCDM method is concerned with two main steps: (i) determination of criteria weights; and (ii) calculating the evaluation of alternatives together with their priority ranking.
\\
\textcolor{blue}{The Technique for Order of Preference by Similarity to Ideal Solution (TOPSIS) is a multiple criteria decision analysis technique being based on the concept that the chosen alternative should have the shortest distance from the positive ideal solution (PIS) as well as the longest distance from the negative ideal solution (NIS). In fact, TOPSIS is an innovative technique for solving MCDM problems with completely unknown weight information about criteria using entropy weight model.}
\\
Suppose that the decision maker is asked to chose one of $m$ alternatives $x_i~(i=1,2,...,m)$ being evaluated based on $n$ criteria $c_j~(j=1,2,...,n)$ by the help of assessment
values in form of PHFSs. Moreover, it is assumed that the criteria are corresponded to the weighting vector $W= (w_1,w_2,...,w_n)$ which are completely unknown with the properties $w_j\geq 0$ and $\sum_{j=1}^{n}w_j=1$.
Then, the probabilistic hesitant fuzzy MCDM technique being based on TOPSIS method \cite{[chen]} is described by the following steps:
\begin{itemize}
  \item \textbf{Step 1.} Determine the weights $w_j$'s corresponding to the criteria $c_j$'s for  $j=1,2,...,n$ by the help of completely unknown information as the following:
\begin{eqnarray}\label{wei-1}
w_j=\frac{1-\overline{Ep}_j}{n-\sum^{n}_{j=1}\overline{Ep}_j},\quad (j=1,2,...,n),
\end{eqnarray}
where 
\begin{eqnarray}\label{en-11}
\overline{Ep}_j=\frac{1}{m}\sum^{m}_{i=1}Ep(h_{j}(x_i)|p_{j}(x_i))=\frac{1}{m}\sum^{m}_{i=1}Ep(h_{ij}|p_{ij}),\quad (j=1,2,...,n).
\end{eqnarray}  
In the above-mentioned equation, the entropy $Ep$ may be replaced by one of the PHFE entropy measures which are proposed in this contribution.
\item \textbf{Step 2.} Calculate the distance measures of alternatives $x_i$'s from
the positive-ideal solution (PIS) $Dp^+(x_i)$, and also from the
negative-ideal solution (NIS) $Dp^-(x_i)$ which are defined as:
\begin{eqnarray}\label{d+}
&&Dp^+(x_i)=\sum_{j=1}^{n}w_j Dp(h_j(x_i)|p_j(x_i),H^+|P^+),\quad (i=1,2,...,m),\\
&&Dp^-(x_i)=\sum_{j=1}^{n}w_j Dp(h_j(x_i)|p_j(x_i),H^-|P^-),\quad (i=1,2,...,m),
\end{eqnarray}
where $Dp$ is given by (\ref{def-dp}), and furthermore, $H^+|P^+ = \{\langle c_j,~h^+_j|p^+_j\rangle~|~c_j\in C\}$ and $H^-|P^- = \{\langle c_j,~h^-_j|p^-_j\rangle~|~c_j\in C\}$ indicate respectively the probabilistic hesitant fuzzy PIS and  the probabilistic hesitant fuzzy NIS in accordance with the benefit and cost criteria. We consider here $h^+_j|p^+_j=\{1|1\}$ and $h^-_j|p^-_j=\{0|1\}$ corresponding to the benefit-index of $j$, and moreover $h^+_j|p^+_j=\{0|1\}$ and $h^-_j|p^-_j=\{1|1\}$ are considered for the cost-index of $j$.
\item \textbf{Step 3.} Calculate the relative closeness of each alternative $x_i$ regarding 
to the ideal solution in the form of
\begin{eqnarray}\label{dd+}
&&Cp(x_i)=\frac{Dp^-(x_i)}{Dp^+(x_i)+Dp^-(x_i)},\quad (i=1,2,...,m).
\end{eqnarray}
\item \textbf{Step 4.} Rank all the alternatives with respect to the increasing order of relative
closeness of $Cp(x_i)$ for $ (i=1,2,...,m)$, that is, the bigger value of $Cp(x_i)$ determines the better alternative $x_i$.
\end{itemize}

Now, we are in a position to consider the problem of 
MCDM with probabilistic
hesitant fuzzy information which was originally investigated by Pang et al. \cite{pang} and Liu et al. \cite{liu} on the basis of probabilistic
linguistic term sets (PLTSs).
\\
In this contribution, in order to transform the PLTS-based practical case concerning strategy initiatives to that of PHFE-based one, we are able to implement the following bijective transformation discussed by  Farhadinia \cite{40ord}.   
Suppose that $S = \{s_t~|~t=0,1,...,2\tau \}$ is a finite and totally ordered discrete linguistic term set (LTS), and $h_S = \{s_t ~|~t\in [0,2\tau] \}$ is a hesitant fuzzy LTS in which $h_\gamma =\{\gamma ~|~\gamma\in [0, 1] \}$ is a HFE. Farhadinia \cite{40ord} stated that the linguistic variable $s_t$ can be transformed into the the membership
degree $\gamma$ by using the function 
\begin{eqnarray*}
\left\{
  \begin{array}{ll}
    g:[0,2\tau]\rightarrow[0,1], & \hbox{} \\
    g(s_t)=\frac{t}{2\tau}=\gamma, & \hbox{}
  \end{array}
\right.
\end{eqnarray*}
and therefore
\begin{eqnarray}\label{trans1}
g(h_S)=\{g(s_t)=\frac{t}{2\tau}~|~t\in[0,2\tau]\}=h_\gamma.
\end{eqnarray}

By the way, following Pang et al. \cite{pang} and Liu et al. \cite{liu}, it is supposed that   
the board of five directors in a company are invited for planing the development
of strategy initiatives during the next five years. To do this end, they are asked to  evaluate three possible projects $x_i~ (i=1,2,3)$ which may be supposed as the strategy initiatives.  In order for comparing the three projects based on their importance degrees, we undertake the following benefit-type four criteria: $c_1$:   
from the  perspective of financial; $c_2$: the satisfaction of customers; $c_3$: from the perspective of internal business process; and $c_4$: from the perspective of learning and growth. We moreover assume that the weight vector of the criteria is completely unknown. To deal with this problem, we implement the TOPSIS method described above.
\\
Pang et al. \cite{pang} and Liu et al. \cite{liu} assumed that the decision makers employ the linguistic term set $S=\{s_0=none,~s_1=very~ low,~s_2=low,~s_3=medium,~s_4=high,~s_5=very~ high,~ s_6=perfect\}$ for evaluating the latter-mentioned projects by the help of PLTSs. 
Here, we transform PLTSs into PHFEs by helping the rule of (\ref{trans1}), and consequently, 
the normalized linguistic decision matrix is transformed to that matrix in the form of PHFEs as given in Table 9.

\begin{center}
{Table 9. The probabilistic hesitant fuzzy decision matrix.}\\
{\small{
\begin{tabular}{|c| c c c c |}
 \hline
 Alternative &  $c_1$  &  $c_2$ & $c_3$  &  $c_4$ \\ \hline
  \hline  
$x_1$ & $\{0.5|0,0.5|0.4,0.66|0.6\}$ & $\{0.33|0,0.33|0.2,0.66|0.8\}$ & $\{0.5|0,0.5|0.2,0.66|0.8\}$ & $\{0.5|0,0.5|0.4,0.83|0.6\}$ \\ \hline
$x_2$ & $\{0.5|0,0.5|0.8,0.83|0.2\}$ & $\{0.33|0.25,0.5|0.5,0.66|0.25\}$ & $\{0.16|0.25,0.33|0.5,0.5|0.25\}$ & $\{0.5|0,0.5|0.8,0.66|0.2\}$ \\ \hline
$x_3$ & $\{0.5|0,0.5|0.6,0.66|0.4\}$ & $\{0.5|0,0.5|0.75,0.66|0.25\}$ & $\{0.5|0.33,0.66|0.33,0.83|0.33\}$ & $\{0.66|0,0.66|0.8,1|0.2\}$ \\ \hline

\end{tabular}
}}\vspace{0.1cm}
\end{center}

Now, by using \textit{Step 1} of the above-mentioned TOPSIS algorithm, we are able to calculate the weight vector of all benefit-type criteria $c_j ~(j=1,2,3,4)$ by the help of all proposed entropy measures. Once again,  let us recall here all the proposed comprehensive entropy measures $Ep_{uvC}$:
\begin{eqnarray*}
&&Ep_{uvC1}(h_A(x)|p_A(x))=\max\{Ep_{uF}(h_A(x)|p_A(x)),Ep_{vNS}(h_A(x)|p_A(x))\};\\
&&Ep_{uvC2}(h_A(x)|p_A(x))=Ep_{uF}(h_A(x)|p_A(x))+Ep_{vNS}(h_A(x)|p_A(x))-Ep_{uF}(h_A(x)|p_A(x))Ep_{vNS}(h_A(x)|p_A(x));\\
&&Ep_{uvC3}(h_A(x)|p_A(x))=\min\{Ep_{uF}(h_A(x)|p_A(x))+Ep_{vNS}(h_A(x)|p_A(x)),1\},
\end{eqnarray*}
for $u=1,2$ and $v=1,2,3$ in which 
\begin{eqnarray*}
&&Ep_{1F}(h(x))=\nonumber \\&&\frac{2}{l_x(l_x+1)}\sum_{i=1}^{l_x}\sum_{j\geq i}
[1-(\frac{1}{3}|1-4h^{\sigma(i)}(x)h^{\sigma(j)}(x)|)^r]\nonumber \\&&
\times[1-(\frac{1}{3}|4(h^{\sigma(i)}(x)+h^{\sigma(j)}(x)-h^{\sigma(i)}(x)h^{\sigma(j)}(x))-3|)^r]
\times \Pi(p_A^{{\sigma(i)}}(x),p_A^{{\sigma(j)}}(x)),\quad r\geq 1;\\
&&Ep_{2F}(h(x))=\nonumber \\&&\frac{2}{l_x(l_x+1)}\sum_{i=1}^{l_x}\sum_{j\geq i}
\frac{2}{3}(\min\{1-2h^{\sigma(i)}(x)h^{\sigma(j)}(x),h^{\sigma(i)}(x)h^{\sigma(j)}(x)\}+1)
\nonumber\\&&
\times
\frac{2}{3}(\min\{2(h^{\sigma(i)}(x)+h^{\sigma(j)}(x)-h^{\sigma(i)}(x)h^{\sigma(j)}(x))-1,\nonumber\\&&
\hspace{4cm} 2-2(h^{\sigma(i)}(x)+h^{\sigma(j)}(x)-h^{\sigma(i)}(x)h^{\sigma(j)}(x))\}+1)
\times \Pi(p_A^{{\sigma(i)}}(x),p_A^{{\sigma(j)}}(x));
\end{eqnarray*}
are the fuzziness-based entropy measures, 
and 
\begin{eqnarray*}
&&Ep_{1NS}(h(x))=\frac{2}{\max\{2,l_x(l_x-1)\}}\sum_{i=1}^{l_x}\sum_{j\geq i}
[\frac{2|h^{\sigma(i)}(x)-h^{\sigma(j)}(x)|}{1+|h^{\sigma(i)}(x)-h^{\sigma(j)}(x)|}]^{\Pi(p_A^{{\sigma(i)}}(x),p_A^{{\sigma(j)}}(x))}; \\
&&Ep_{2NS}(h(x))=\frac{2}{\max\{2,l_x(l_x-1)\}}\sum_{i=1}^{l_x}\sum_{j\geq i}
[\frac{log(1+|h^{\sigma(i)}(x)-h^{\sigma(j)}(x)|)}{log(2)}]^{\Pi(p_A^{{\sigma(i)}}(x),p_A^{{\sigma(j)}}(x))};
\\
&&Ep_{3NS}(h(x))=\frac{2}{\max\{2,l_x(l_x-1)\}}\sum_{i=1}^{l_x}\sum_{j\geq i}
[|h^{\sigma(i)}(x)-h^{\sigma(j)}(x)|e^{|h^{\sigma(i)}(x)-h^{\sigma(j)}(x)|-1}]^{\Pi(p_A^{{\sigma(i)}}(x),p_A^{{\sigma(j)}}(x))},
\end{eqnarray*}
denote the non-specificity-based entropy measures.\\
Now, in accordance with the above-mentioned entropy measures, we represent the corresponding results together with Pang et al.'s \cite{pang} and Liu et al.'s \cite{liu} weight vectors in the summarized form which are all shown  in Table 10.

\newpage
\begin{center}
{Table 10. Criteria weights $w_i~(i = 1, 2, 3, 4)$ and their corresponding ranking orders.}
{\scriptsize{
\begin{tabular}{|c| c c c c |}
 \hline
 Method &  $w_1$  &  $w_2$ & $w_3$  &  $w_4$ \\ 
 & & Rankings  & & \\ \hline
  \hline  
Pang et al.'s \cite{pang} technique & $0.138$ & $0.304$ & $0.416$ & $0.142$ \\
& $w_3$ & $>w_2$ & $>w_4$ &$>w_1$ \\ \hline
Liu et al.'s \cite{liu} technique & $0.1736$ & $0.2425$ & $0.4022$ & $0.1817$ \\
 & $w_3$ & $>w_2$ & $>w_4$ &$>w_1$ \\ \hline
 \hline
 
Proposed technique based on $Ep_{11C1}$ &     0.2535&    0.2505&    0.5223 &   0.2509  \\
 & $w_3$ & $>w_1$ & $>w_4$ &$>w_2$ \\ \hline
Proposed technique based on $Ep_{12C1}$ & 0.2535&    0.2505 &   0.5115 &   0.2509 \\
 & $w_3$ & $>w_1$ & $>w_4$ &$>w_2$ \\ \hline
Proposed technique based on $Ep_{13C1}$ & 0.2535&    0.2505&    0.4843&    0.2509 \\
& $w_3$ & $>w_1$ & $>w_4$ &$>w_2$ \\ \hline

Proposed technique based on $Ep_{21C1}$ &    0.2731&    0.2389 &   0.5223&    0.2633  \\
 & $w_3$ & $>w_1$ & $>w_4$ &$>w_2$ \\ \hline
Proposed technique based on $Ep_{22C1}$ & 0.2731 &   0.2389&    0.5115 &   0.2633 \\
  & $w_3$ & $>w_1$ & $>w_4$ &$>w_2$ \\ \hline
Proposed technique based on $Ep_{23C1}$ & 0.2731&    0.2470 &   0.4843&    0.2633 \\
  & $w_3$ & $>w_1$ & $>w_4$ &$>w_2$ \\ \hline
 \hline

Proposed technique based on $Ep_{11C2}$ &    0.4200&    0.4153&    0.6394&    0.2770  \\
  & $w_3$ & $>w_1$ & $>w_2$ &$>w_4$ \\ \hline
Proposed technique based on $Ep_{12C2}$ &0.4172&    0.4210 &   0.6312&    0.2823 \\
   & $w_3$ & $>w_2$ & $>w_1$ &$>w_4$ \\ \hline
Proposed technique based on $Ep_{13C2}$ &     0.4105&    0.4357&    0.6106&    0.2947  \\
  & $w_3$ & $>w_2$ & $>w_1$ &$>w_4$ \\ \hline

Proposed technique based on $Ep_{21C2}$ &   0.4352&    0.4062&    0.6297&    0.2890  \\
  & $w_3$ & $>w_1$ & $>w_2$ &$>w_4$ \\ \hline
Proposed technique based on $Ep_{22C2}$ & 0.4325&    0.4119&    0.6213 &   0.2941 \\
& $w_3$ & $>w_1$ & $>w_2$ &$>w_4$ \\ \hline
Proposed technique based on $Ep_{23C2}$ & 0.4259 &   0.4269&    0.6002&    0.3064 \\
& $w_3$ & $>w_2$ & $>w_1$ &$>w_4$ \\ \hline
 \hline

Proposed technique based on $Ep_{11C3}$ &  0.4765&    0.4704&    0.7673&    0.2858  \\
& $w_3$ & $>w_1$ & $>w_2$ &$>w_4$ \\ \hline
Proposed technique based on $Ep_{12C3}$ &0.4728&    0.4779 &   0.7565 &   0.2928 \\
& $w_3$ & $>w_2$ & $>w_1$ &$>w_4$ \\ \hline
Proposed technique based on $Ep_{13C3}$ &  0.4638&    0.4975&    0.7292&    0.3094  \\
& $w_3$ & $>w_2$ & $>w_1$ &$>w_4$ \\ \hline

Proposed technique based on $Ep_{21C3}$ &  0.4961&    0.4587&    0.7471&    0.2982  \\
& $w_3$ & $>w_1$ & $>w_2$ &$>w_4$ \\ \hline
Proposed technique based on $Ep_{22C3}$ & 0.4924&    0.4663&    0.7362 &   0.3051 \\
& $w_3$ & $>w_1$ & $>w_2$ &$>w_4$ \\ \hline
Proposed technique based on $Ep_{23C3}$ & 0.4834&    0.4859&    0.7090&    0.3218 \\
& $w_3$ & $>w_2$ & $>w_1$ &$>w_4$ \\ \hline
 
\end{tabular}
}}
\vspace{0.1cm}
\end{center}

As can be observed from Table 10, although, the ranking of criteria weights according to Pang et al.'s \cite{pang} and Liu et al.'s \cite{liu} entropy measures and those proposed in this contribution are not exactly the same, but they all return $c_3$ as the most appropriate criteria.

However, since the main purpose of this study is to show the merits of proposed entropy measures for PHFEs and moreover due to lack of space, we prefer here not to dwell on the details of the TOPSIS procedure. Hence,  by summarizing all the next steps of TOPSIS technique, we just represent the ranking orders of alternatives in accordance with some entropy measures in Table 11. 

\begin{center}
{Table 11. Comparison of different methods.}

{\scriptsize{
\begin{tabular}{|c| c |}
 \hline
 Method &  Ranking of alternatives  \\ 
   \hline  
Pang et al.'s \cite{pang} TOPSIS-based technique & $x_1>x_3>x_2$ \\
& $(0.0000, -1.8000, -0.6000)$ \\ \hline 
Pang et al.'s \cite{pang} aggregation-based technique & $x_1>x_3>x_2$ \\
& \textcolor{blue}{$(1.2600,    0.8900,    1.2400)$} \\ \hline

Liu et al.'s \cite{liu} technique & $x_1>x_3>x_2$ \\ 
& $(0.4737 ,0.3379 , 0.4733)$ \\ \hline 
 \hline
 
TOPSIS technique based on $Ep_{11C1}$ &  $x_1>x_3>x_2$  \\
 & $(0.5061,    0.0003,    0.0400)$ \\ \hline 
%
%
TOPSIS technique based on $Ep_{22C1}$ &  $x_1>x_3>x_2$  \\
 & $(0.5826,    0.1863,    0.2294)$ \\ \hline 
\hline

TOPSIS technique based on $Ep_{11C2}$ &   $x_1>x_3>x_2$  \\
 & $(0.5044,    0.0043,    0.0361)$ \\ \hline 
%
%
TOPSIS technique based on $Ep_{22C2}$ &   $x_1>x_2>x_3$  \\
 & $(0.5584,    0.3055,    0.2020)$ \\ \hline
\hline 
 
TOPSIS technique based on $Ep_{11C3}$ &   $x_1>x_3>x_2$  \\
 & $(0.5043,    0.0031,    0.0351 )$ \\ \hline
%
%
TOPSIS technique based on $Ep_{22C3}$ &   $x_1>x_2>x_3$  \\
 & $(0.5569,    0.2538,    0.2046)$ \\ \hline
 
\end{tabular}
}}
\vspace{0.1cm}
\end{center}

From Table 11, we find that most of the ranking results of alternatives based on 
Pang et al.'s \cite{pang} extended TOPSIS and aggregation techniques, Liu et al.'s \cite{liu} TOPSIS technique and the proposed ones are the same except for the two results corresponding to $Ep_{22C2}$ and $Ep_{22C3}$. What needs to be mentioned here is that using Pang et al.'s \cite{pang} technique, the criteria weights are computed by maximizing the deviation
of weighted assessments for all alternatives
under each criterion, meanwhile, both Liu et al.'s \cite{liu} technique and the proposed one implement entropy measure of assessments under each criterion in order to compute the criteria weights.
Furthermore, the difference between Liu et al.'s \cite{liu} entropy measures and the proposed ones is that the latter measures return the entropy of PHFEs while the former return the entropy of PLTSs.

\textcolor{red}{To conclude this contribution let us here summarize the main conclusions of the presented work:
\begin{itemize}
\item As demonstrated in Section \ref{sec4}, the comparative analysis between the performance of well-established entropy measures for PHFEs and the only existing PHFE entropy measures of Su et al. \cite{su} indicate that both proposed fuzziness- and 
non-specific-based entropy measures
produce much more logical results than Su et al's \cite{su} entropy measures.
\item In view of the results given in Section \ref{sec5} regarding to the considered MCDM techniques including those of Pang et al. \cite{pang}, Liu et al. \cite{liu} and the proposed one, we observed that the rankings of alternatives are mostly the same.
This is while,
Pang et al.'s \cite{pang} and Liu et al.'s \cite{liu} entropy measures are restricted to PLTS values, and the proposed entropy measures return the entropy amount of extended form of PLTSs, that is, PHFEs.
\end{itemize}
}

\section{Conclusions and further research perspectives}

In this contribution, we  critically reviewed the only class of existing entropy measures for PHFEs by emphasising on their improper and inconsistent use. 
Then, we suggested a fresh-full axiomatic framework of entropy
measures for PHFEs by taking fully into account two facets of uncertainty associated with PHFEs,  known as fuzziness and non-specificity. 
Corresponding to fuzziness and non-specificity aspects, we generalized the technique of constructing the entropy measures for PHFEs, 
{and also we developed the class of entropy measures for PHFEs by introducing PHFE entropy-based distance measures. Then, we 
applied both concepts of PHFE entropy measures and PHFE entropy-based distance measures}
into the
decision making context such as the strategy initiatives.  
The comparison results verified that the proposed measures of entropy for PHFEs do  capture successfully the
intrinsic characteristics of PHFEs and produce valid results.
\textcolor{blue}{However, it needs to be pointed out that how to measure uncertainty is still an open issue. We introduced here the two aspects of uncertainty related to a PHFE together with their combination, and of course these aspects are only two expressions of uncertainty measure. There are other expressions that should be studied more in-depth, and it needs to be investigated whether those expressions are at all applicable under realistic conditions or not. Thus it is noteworthy that although the proposed entropy measures are helpful for evaluating uncertainty information in decision making, further studies is required to explore their applications in real-life domains.
Hence, the future work could be extended to the development of application for PHFE entropy
measures in different areas of decision making,
\textcolor{red}{especially, those are based on the relationship of entropy measure with distance measure \cite{dis-xiao}, divergence measure \cite{div-xiao} and belief entropy measure \cite{bel-xiao}.}
Moreover, there exist good prospects for the further study of the proposed kinds of entropy measures which can be extended to the other fuzzy sets including neutrosophic sets \cite{kum}, picture fuzzy sets \cite{josh}, Pythagorean fuzzy sets \cite{sark,wan-li}, Pythagorean probabilistic hesitant fuzzy sets \cite{bato}, q-rung orthopair fuzzy sets \cite{verm}, etc.}




\begin{thebibliography}{0}
{\small{


\bibitem{bato} \textcolor{blue}{B. Batool, M. Ahmad, S. Abdullah, S. Ashraf, R. Chinram, 
Entropy based Pythagorean probabilistic hesitant fuzzy decision making technique and its application for fog-haze factor assessment problem, Entropy 22 (2020) 318.}

\bibitem{[chen]} C.T. Chen, Extensions of the TOPSIS for group decision-making under fuzzy
environment, Fuzzy Sets Syst. 114 (2000) 1-9.


\bibitem{ding} J. Ding, Z.S. Xu, N. Zhao, An interactive approach to probabilistic
hesitant fuzzy multi-attribute group decision
making with incomplete weight information, J. Intell. Fuzzy Syst. 32 (2017) 2523-2536.

\bibitem{[5EntM]} B. Farhadinia, Information measures for hesitant
fuzzy sets and interval-valued hesitant fuzzy sets,  Info.
Sci., 10 (2013) 129-144. 

\bibitem{[38tfar]} B. Farhadinia, A multiple criteria decision making model with entropy
weight in an interval-transformed hesitant fuzzy environment, Cogn. Comput. 9 (2017) 513-525.

\bibitem{40ord} B. Farhadinia, Ordered weighted hesitant fuzzy information fusion-based
approach to multiple attribute decision making with probabilistic
linguistic term sets, Fundamenta Info. 159 (2018) 361-383.

\bibitem{fardet} B. Farhadinia, Determination of entropy measures for the ordinal scale-based linguistic models, Info. Sci. 369 (2016) 63-79.

\bibitem{farxu2} B. Farhadinia, Z. Xu, Novel hesitant fuzzy linguistic entropy and cross-entropy measures in multiple criteria decision making, App. Intell. 48 (2018) 3915-3927.

\bibitem{kum} \textcolor{blue}{A. Kumar, C.P. Gandhi, Y. Zhou, H. Tang, J. Xiang, 
Fault diagnosis of rolling element bearing based on symmetric cross entropy of neutrosophic sets, 
Measurement 152 (2020) 107318.}

\bibitem{liu} H. Liu, L. Jiang, Z. Xu, Entropy measures of probabilistic linguistic term sets, Int. J. Comput. Intell. Syst. 11 (2018) 45-57. 

\bibitem{josh} \textcolor{blue}{R. Joshi, A new picture fuzzy information measure based on Tsallis-Havrda-Charvat concept with applications in presaging poll outcome, Comp. Appl. Math. 39 (2020) https://doi.org/10.1007/s40314-020-1106-z.}

\bibitem{pal} \textcolor{blue}{N.R. Pal, H. Bustince, M. Pagola, U.K. Mukherjee, D.P. Goswami, G. Beliakov, Uncertainties with Atanassov's intuitionistic fuzzy sets: fuzziness and lack of knowledge, Inf. Sci. 228 (2013) 61-74.}  

\bibitem{pang} Q. Pang, H. Wang, Z. Xu, Probabilistic linguistic term sets in
multi-attribute group decision making, Info. Sci. 369 (2016) 128-143.

\bibitem{sark} \textcolor{blue}{B. Sarkar, A. Biswas, A unified method for Pythagorean fuzzy multicriteria group decision-making using entropy measure, linear programming and extended technique for ordering preference by similarity to ideal solution, Soft Comput 24 (2020) 5333-5344.}

\bibitem{[anov]} C. Song, Z.S. Xu, H. Zhao, A novel comparison of probabilistic hesitant fuzzy
elements in multi-criteria decision making, Symmetry 177 (2018) doi:10.3390/sym10050177. 

\bibitem{su} Z. Su, Z.S. Xu, H. Zhao, Z. Hao, B. Chen, Entropy measures for probabilistic hesitant
fuzzy information, IEEE Access (2019) doi:10.1109/ACCESS.2019.2916564.


\bibitem{[29tor]} V. Torra, Hesitant fuzzy sets, Int. J. Intell.
Syst. 25 (2010) 529-539.

\bibitem{verm} \textcolor{blue}{R. Verma, Multiple attribute group decision-making based on order-a divergence and entropy measures under q-rung orthopair fuzzy environment, Int. J. Intell. Syst.  35 (2020) 718-750. } 


\bibitem{wan-li} \textcolor{blue}{L. Wang, N. Li, Pythagorean fuzzy interaction power Bonferroni mean aggregation operators in multiple attribute decision making, Int. J. Intell. Syst. 35 (2020) 150-183.}

\bibitem{[21EntM]} C. Wei, F. Yan, R.M. Rodriguez, Entropy measures
for hesitant fuzzy sets and their application in
multi-criteria decision-making,  J. Intell. Fuzzy Syst. 31 (2016) 673-685.

\bibitem{[6xia]} M. Xia, Z. Xu, Hesitant fuzzy information aggregation in decision
making, Int. J. Approx. Reasoning 52 (2011) 395-407.


\bibitem{dis-xiao} \textcolor{red}{F. Xiao, A distance measure for intuitionistic fuzzy
sets and its application to pattern
classification problems, IEEE Trans Syst Man Cybern Syst. (2019) https://doi.org/10.1109/TSMC.2019.2958635.}

\bibitem{div-xiao} \textcolor{red}{F. Xiao, A new divergence measure for belief functions in D-S evidence theory for multisensor data fusion, Inf Sci 514 (2020) 462-483.}

\bibitem{bel-xiao} \textcolor{red}{F. Xiao, EFMCDM: evidential fuzzy multicriteria decision making based on belief entropy, IEEE Trans Fuzzy Syst.  (2019) https://doi.org/10.1109/TFUZZ.2019.2936368.}

\bibitem{[Exm1]} Z. Xu, M. Xia, Hesitant fuzzy entropy and cross-entropy and
their use in multiattribute decision-making, Int. J. Intell. Syst. 27 (2012) 799-822.


\bibitem{[8xu]} Z. Xu, W. Zhou, Consensus building with a group of decision makers
under the hesitant probabilistic fuzzy environment,' Fuzzy Optim. Decis.
Making 16 (2017) 481-503.

\bibitem{[31yue]} L. Yue, M. Sun, Z. Shao, The probabilistic hesitant
fuzzy weighted average operators and their application in
strategic decision making, J. Info. Comput. Sci. 10 (2013) 3841-3848.

\bibitem{[32zen]} S.Z. Zeng, J.M. Merigo, W. Su, The uncertain probabilistic
OWA distance operator and its application in
group decision making, App. Math. Model. 37 (2013) 6266-6275.

\bibitem{[15anov]} S. Zhang, Z.S. Xu, Y. He, Operations and integrations of probabilistic hesitant fuzzy information in decision
making,  Info. Fusion 38 (2017) 1-11. 



\bibitem{[30zhu]} B. Zhu, Decision method for research and application based
on preference relation, Nanjing: Southeast University, 2014.





 




}}
 \end{thebibliography}
\end{document}